\documentclass[10pt,twocolumn,letterpaper]{article}

\usepackage{cvpr}              

\usepackage{graphicx}
\usepackage{amsmath}
\usepackage{amssymb}
\usepackage{booktabs}
\usepackage{multirow}
\usepackage{tabularx}
\usepackage[ruled]{algorithm2e}
\usepackage[accsupp]{axessibility}  
\usepackage{caption}
\usepackage{colortbl}
\usepackage{soul} 
\usepackage{color, xcolor} 
\definecolor{mygray}{gray}{0.95}
\usepackage{subcaption}
\usepackage[pagebackref,breaklinks,colorlinks]{hyperref}

\usepackage[capitalize]{cleveref}
\crefname{section}{Sec.}{Secs.}
\Crefname{section}{Section}{Sections}
\Crefname{table}{Table}{Tables}
\crefname{table}{Tab.}{Tabs.}

\begin{document}


\title{SNP-S$^3$: Shared Network Pre-training and Significant Semantic Strengthening for Various Video-Text Tasks}

\author{Xingning Dong\textsuperscript{\rm 1}{\footnotemark[1]}, \quad Qingpei Guo\textsuperscript{\rm 1}{\footnotemark[1]}, \quad Tian Gan\textsuperscript{\rm 2}{\footnotemark[2]}, \quad Qing Wang\textsuperscript{\rm 1}, 
\\
Jianlong Wu\textsuperscript{\rm 3}, \quad Xiangyuan Ren\textsuperscript{\rm 1}, \quad Yuan Cheng\textsuperscript{\rm 4}, \quad Wei Chu\textsuperscript{\rm 1} 
\\
\normalsize{\textsuperscript{\rm 1}Ant Group, \quad \textsuperscript{\rm 2}Shandong University, \quad \textsuperscript{\rm 3}Harbin Institute of Technology (Shenzhen), \quad \textsuperscript{\rm 4}Fudan University}
\\
{\tt\small dongxingning1998@gmail.com, \quad qingpei.gqp@antgroup.com, \quad gantian@sdu.edu.cn,}
\\
{\tt\small wq176625@antgroup.com, \quad jlwu1992@pku.edu.cn, \quad xiangyuan.rxy@antgroup.com,}
\\
{\tt\small cheng$\_$yuan@fudan.edu.cn, \quad weichu.cw@antgroup.com}
}


\maketitle
\renewcommand{\thefootnote}{\fnsymbol{footnote}}
\footnotetext[1]{Xingning Dong and Qingpei Guo contributed equally to this manuscript.}
\footnotetext[2]{Tian Gan is the corresponding author.}

\begin{abstract}

We present a framework for learning cross-modal video representations by directly pre-training on raw data to facilitate various downstream video-text tasks. Our main contributions lie in the pre-training framework and proxy tasks. First, based on the shortcomings of two mainstream pixel-level pre-training architectures (limited applications or less efficient), we propose Shared Network Pre-training (SNP). By employing one shared BERT-type network to refine textual and cross-modal features simultaneously, SNP is lightweight and could support various downstream applications. Second, based on the intuition that people always pay attention to several ``significant words'' when understanding a sentence, we propose the Significant Semantic Strengthening (S$^{3}$) strategy, which includes a novel masking and matching proxy task to promote the pre-training performance. Experiments conducted on three downstream video-text tasks and six datasets demonstrate that, we establish a new state-of-the-art in pixel-level video-text pre-training; we also achieve a satisfactory balance between the pre-training efficiency and the fine-tuning performance. The codebase are available at \href{https://github.com/alipay/Ant-Multi-Modal-Framework/tree/main/prj/snps3_vtp}{https://github.com/alipay/Ant-Multi-Modal-Framework/tree/main/prj/snps3$\_$vtp}.

\end{abstract}

\section{Introduction}
\label{sec:intro}

Owing to successful applications of pre-training methods in NLP \cite{cui2021pre, 10.1145/3566126} and CV \cite{9748114, 9337201}, more and more researchers attempt to explore this ``Pre-training $\&$ Fine-tuning'' paradigm in the video-text field \cite{10.1145/3473140, 10.1145/3585388}, which has achieved remarkable performance gain in various downstream video understanding tasks, such as video-text retrieval \cite{9709794, 8302915, 9463746}, video question answering \cite{9097301, 9311693, 4633656}, and video reasoning \cite{grunde2021agqa, yi2019clevrer, wu2021star, chen2022comphy, zellers2022merlot}. There are two mainstream paradigms in current video-text pre-training methods: the feature-level paradigm and the pixel-level one.

\begin{figure}[t]
	\centering
    	\begin{subfigure}{0.265\linewidth}
    		\includegraphics[width=1.0\textwidth]{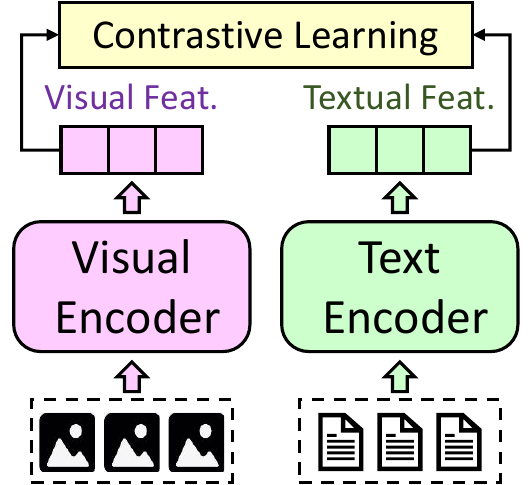}
    		\caption{Twin-tower}
                \label{intro-ant-a}
    	\end{subfigure}
    	\begin{subfigure}{0.265\linewidth}
    		\includegraphics[width=1.0\textwidth]{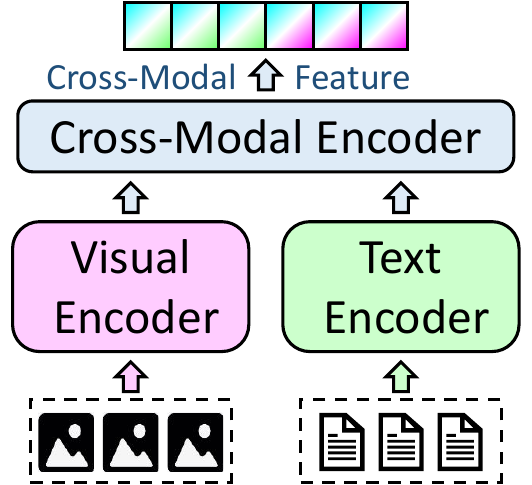}
    		\caption{Three-fusion}
                \label{intro-ant-b}
    	\end{subfigure}
    	\begin{subfigure}{0.435\linewidth}
    		\includegraphics[width=1.0\textwidth]{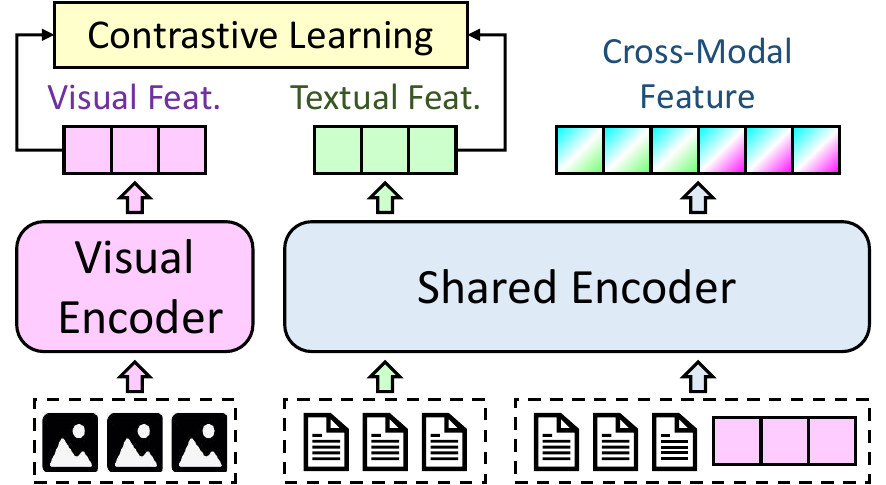}
    		\caption{The proposed \textbf{SNP} paradigm}
                \label{intro-ant-c}
    	\end{subfigure}
    	
	\caption{Comparison of mainstream pixel-level pre-training architectures: a) Twin-tower-based, b) Three-fusion-based, and c) the proposed Shared Network Pre-training (\textbf{SNP}) methods.}
    \label{intro-ant}
\end{figure}

Compared with feature-level pre-training methods \cite{xu2021vlm, li2020hero, luo2021coco} that employ off-the-shelf visual and textual features extracted by frozen models, pixel-level pre-training methods \cite{lei2021less, ge2022bridging, bain2021frozen} treat raw visual pixels and text tokens as inputs, which could optimize the cross-modal learning ability in an end-to-end manner. Thus, the pixel-level paradigm tends to achieve better performance and has been widely followed. There are two mainstream pixel-level pre-training architectures, \textit{i}.\textit{e}., twin-tower-based \cite{ge2022bridging, fang2021clip2video, bain2021frozen} in Figure \ref{intro-ant-a} and three-fusion-based \cite{lei2021less, li2022align, fu2021violet} in Figure \ref{intro-ant-b}. Twin-tower-based models are usually lightweight and time-efficient; however, since they do not generate cross-modal video representations, their applications are limited mainly in the cross-modal retrieval task. Three-fusion-based models usually contain three separate encoders to embed visual, textual, and cross-modal features. Though they could support various downstream video understanding applications, they usually contain massive training parameters, leading to computational inefficiency and high cost of GPU memory. 

Towards this end, we propose a new architecture that not only supports various video-text tasks like three-fusion-based models, but also as lightweight as twin-tower-based ones. Based on the thorough investigation, we observe that: 1) the text and cross-modal encoder in conventional three-fusion-based models \cite{lei2021less, fu2021violet, li2022align} are mainly BERT-type transformers; 2) the inputs and outputs of both encoders are token-type features; and 3) as CLIP4Clip \cite{luo2021clip4clip} pointed out, it is hard to find suitable parameters to initialize the cross-modal encoder, leading to sub-optimal pre-training performance. Therefore, we propose the Shared Network Pre-training (\textbf{SNP}) method. As shown in Figure \ref{intro-ant-c}, \textbf{SNP} employs a shared BERT-type network to refine textual and cross-modal features simultaneously, combining the advantages of twin-tower-based and three-fusion-based methods.

In order to promote the cross-modal interaction for better performance, current video-text pre-trained models would set several proxy tasks, where Masked Language Modeling (MLM) and Global Vision-Text Matching (GVTM) are two widely-employed ones. However, for the conventional MLM, some masking words could be easily filled according to grammar without reviewing the image. \textit{E}.\textit{g}., given a sentence ``[?] \textit{boy} \textit{in} \textit{red} [?] \textit{sitting} [?] a \textit{skateboard}'', one can directly refer to ``a'', ``is'' and ``on''. Thus, it seems that conventional MLM could hardly benefit the cross-modal interaction. For GVTM that aims to model the cross-modal alignment, current methods usually take pair-wised visual features and global-pooling textual features (refer to the hidden state of the token [\textit{cls}]) as inputs. However, global-pooling textual features focus on the sentence level, which would omit the local information of some informative semantics at the word level, leading to limited performance.

\begin{figure}[t]
	\centering
	\includegraphics[width=0.49\textwidth]{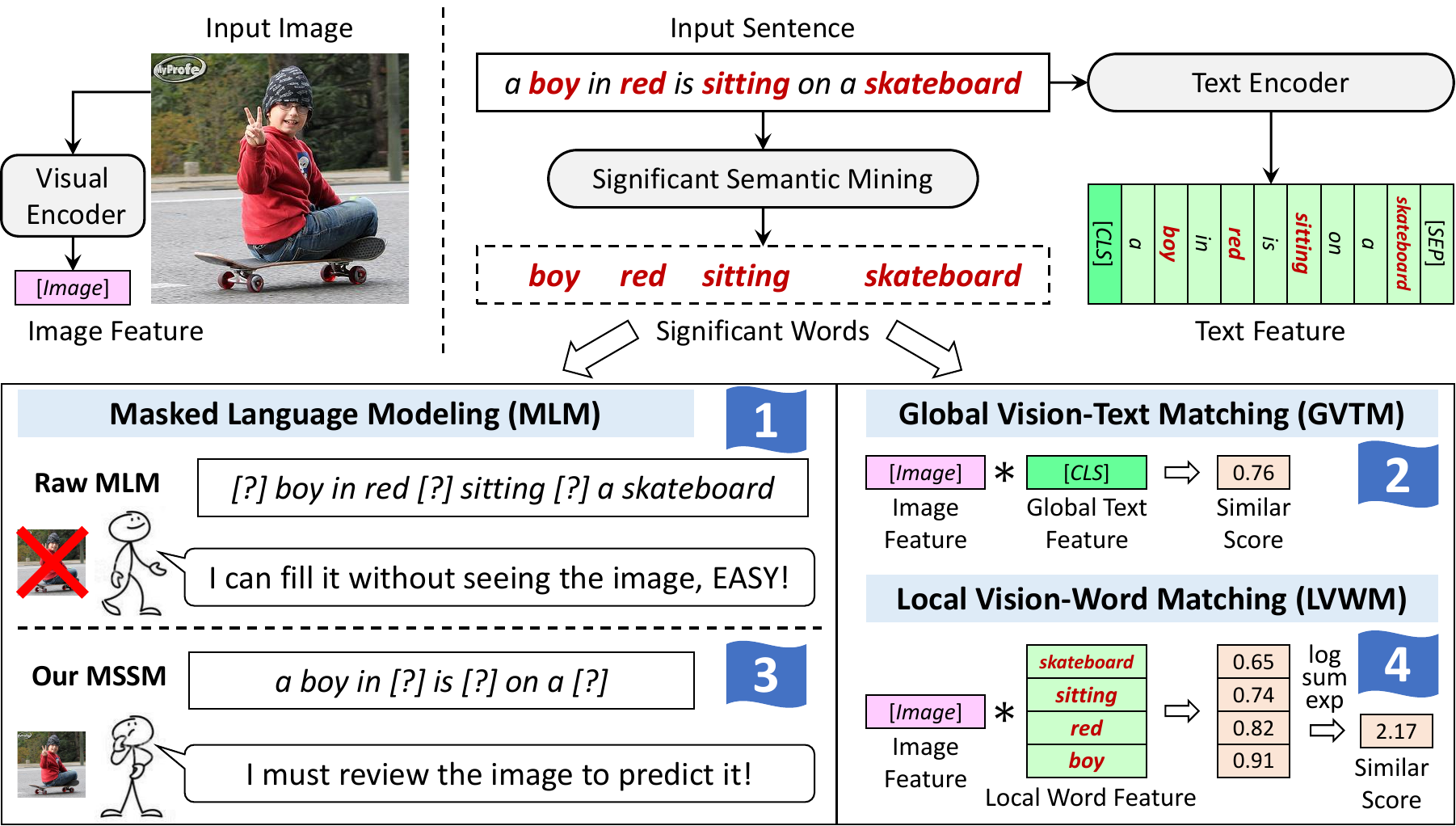}
	\caption{Comparison of two widely-employed masking and matching proxy tasks (MLM-1 and GVTM-2) and our improved version (MSSM-3 and LVWM-4).}
	\label{intro-sss}
\end{figure}
Intuitively, humans usually capture several ``significant words'' when understanding a sentence. \textit{I}.\textit{e}., some words (\textit{e}.\textit{g}., verbs and nouns) would provide significant information while others (\textit{e}.\textit{g}., prepositions and conjunctions) only play the role of ``lubricants'' to make the sentence fluent and vivid. In order to improve the cross-modal interaction, we hope pre-trained models would lay their emphasis on those significant words rather than other trivial ones. Therefore, we propose the Significant Semantic Strengthening (${\textbf{S}}^{\textbf{3}}$) strategy, leading pre-trained models to automatically find and emphasize these significant semantics within the input sentence. As shown in the third and fourth parts of Figure \ref{intro-sss}, ${\textbf{S}}^{\textbf{3}}$ includes two novel proxy tasks for better cross-modal interaction: 1) Masked Significant Semantic Modeling (MSSM) masks out informative words to force models to resume these clozes from textual and visual information, replacing the conventional MLM. 2) Local Vision-Word Matching (LVWM) learns the cross-modal interaction at the word level, which is a complementary to existing GVTM. 

Our contributions are summarized in three-folds:

\begin{itemize}

	\item We propose Shared Network Pre-training (SNP), which is a lightweight pixel-level pre-training method and could support various downstream video-text applications.
	
	\item 
	We propose the Significant Semantic Strengthening (\textbf{S}$^{\textbf{3}}$) strategy, including two novel proxy tasks (MSSM and LVTM), which is model-agnostic, parameter-free, and could facilitate the cross-modal interaction.
	
	\item Experiments conducted on three downstream video-text tasks and six datasets indicate the superiority of our proposed method, which establishes a new state-of-the-art in the field of video-text pre-training.

\end{itemize}


\section{Related Work}
\label{sec:related}

\noindent\textbf{Video-Text Pre-training and Fine-Tuning.} \\
Inspired by superior performance of Transformers \cite{9751606, 9770033, 9895256} and BERT \cite{devlin2018bert, huang2020pixel, 10013737}, video-text pre-training has attracted increasing interest in recent years, which could be roughly divided into feature-level pre-training methods \cite{xu2021vlm, luo2020univl, li2020hero} and pixel-level ones \cite{zhu2020actbert, lei2021less, bain2021frozen}. Since the former approaches employ offline visual and textual features extracted from frozen models(\textit{e}.\textit{g}., S3D \cite{xie2018rethinking} and DistillBert \cite{sanh2019distilbert}), they would limit the fine-tuning performance as there remain domain gaps between pre-training datasets and frozen feature extractors. While for pixel-level video pre-training methods, they attempt to learn cross-modal representations from raw visual pixels and text tokens in an end-to-end manner, whose frameworks are mainly twin-tower-based \cite{bain2021frozen, fang2021clip2video, wang2021object} architectures or three-fusion-based \cite{lei2021less, li2022align, fu2021violet} ones. However, both architectures have their limitations: Twin-tower-based methods could hardly support various downstream video understanding tasks like the latter. In contrast, three-fusion-based methods contain more parameters and are not as lightweight as the former. Due to the high cost of GPU memory, conventional three-fusion-based methods mainly pre-train their models on large-scale image-text datasets, and fine-tune them on downstream video-text tasks.

In this work, we propose Shared Network Pre-training, which is a lightweight pixel-level pre-training architecture and could support various downstream video-text tasks.

\noindent\textbf{Significant Element Mining in Video Pre-Training.} \\ Recently a few video-text pre-training methods have started to consciously recognize and emphasize the ``significant elements'' in various proxy tasks for better fine-tuning performance. 1) For \textbf{masking} tasks, MERLOT \cite{zellers2021merlot} and VIOLET \cite{fu2021violet} propose the Attended Masking (AM) strategy, which optimizes conventional MLM by masking out 50$\%$ of tokens with high attention weights calculated from a language-only transformer (MERLOT) or a cross-modal encoder (VIOLET). Different from those methods, we optimize conventional MLM by explicitly leveraging the Parts-of-Speech (POS) tags within a sentence. 2) For \textbf{matching} tasks, TACo \cite{yang2021taco} proposes a token-level contrastive loss based on the maximum dot-products of visual and textual token features. However, TACo may still omit some local information as it only takes one token embedding into computation like conventional GVTM. Besides, 3) some innovative proxy tasks that leverage those significant elements have been proposed, \textit{e}.\textit{g}., Multiple Choice Questions (MCQ) \cite{ge2022bridging} first builds several questions by erasing verb/noun phrases of a sentence. It then forces the model to select right answers from several candidates. 

Based on these successful explorations, we attempt to leverage the ``Significant Elements'' for better cross-modal interaction in both masking and matching proxy tasks.


\section{Methodology}

We propose \textbf{SNP}-$\textbf{S}^\textbf{3}$, which is a pixel-level video pre-training method following the conventional protocol that first pre-trains on large-scale image-text datasets and then fine-tunes on downstream video-text tasks. Moreover, we report the results pre-trained on video-text datasets in the EXPERIMENT section for fair comparison. To achieve a satisfactory balance between the pre-training efficiency and the fine-tuning performance, we simplify the conventional framework and propose two novel proxy tasks.

\subsection{Pre-training on Image-Text Datasets} 

Given a mini-batch (denoted as $\mathcal{B}$) of images $\{I_{i}\}_{i=1}^\mathcal{|B|}$ and their corresponding descriptions $\{S_{i}\}_{i=1}^\mathcal{|B|}$, pixel-level pre-training methods would first extract visual features $\{\mathbf{v}_{i}\}_{i=1}^\mathcal{|B|}$ and textual features $\{\mathbf{t}^{cls}_{i}\}_{i=1}^\mathcal{|B|}$ from raw data, and then generate cross-modal video representations $\{\mathbf{m}^{cls}_{i}\}_{i=1}^\mathcal{|B|}$ to facilitate various downstream video-text tasks (Note that twin-tower-based methods would skip this step).

\subsubsection{Three-fusion-based Pre-training Architecture}

As shown in Figure \ref{intro-ant-b}, conventional three-fusion-based pre-training methods usually contain a visual encoder ${E}_{vis}$, a text encoder ${E}_{txt}$, and a cross-modal encoder ${E}_{mul}$. Given an image-text pair ($I$, $S$), we first employ a BERT embedder ${E}_{B}$ to process the sentence $S$ into fixed-length token embeddings $\mathbf{W} = [{\mathbf{w}}^{cls}, {\mathbf{w}}^{1}, {\mathbf{w}}^{2}, \cdots, {\mathbf{w}}^{{N}_{t}-1}]$, where $\mathbf{W} \in {\mathbb{R}}^{{N}_{t}*d}$, ${N}_{t}$ is the length of tokens, and $d$ is the embedding dimension. Notably, each element in $\mathbf{W}$ except the special tokens (\textit{e}.\textit{g}., [$cls$]) could be treated as a word embedding. We then employ ${E}_{vis}$ and ${E}_{txt}$ to obtain image features $\mathbf{v} \in {\mathbb{R}}^{1*d}$ and textual features $\mathbf{T} \in {\mathbb{R}}^{{N}_{t}*d}$ from raw image pixels $I$ and token embeddings $\mathbf{W}$. Afterwards, we concatenate these two features into [$\mathbf{T}$, $\mathbf{v}$], and feed them into ${E}_{mul}$ to obtain cross-modal features $\mathbf{M} \in {\mathbb{R}}^{({N}_{t}+1)*d}$. This forward process could be formulated as follows: 
\begin{equation}
	\mathbf{M} = {E}_{mul}([E_{txt}({E}_{B}(S)), E_{vis}(I)]),
\end{equation}
where $\mathbf{M} = [{\mathbf{m}}^{cls}, {\mathbf{m}}^{1}, {\mathbf{m}}^{2}, \cdots, {\mathbf{m}}^{{N}_{t}}]$ and $[\cdot , \cdot]$ denotes the concatenation operation. Notably, we treat the first [$cls$] features $\mathbf{m}^{cls}$ as global-pooling video representations.

\subsubsection{SNP: Shared Network Pre-training}

Based on the thorough investigation of twin-tower-based (Figure \ref{intro-ant-a}) and three-fusion-based (Figure \ref{intro-ant-b}) architectures, we aim to absorb their advantages and overcome their shortcomings for better pre-training performance. Therefore, we propose the novel Shared Network Pre-training (SNP) architecture. As illustrated in Figure \ref{intro-ant-c}, we simplify the three-fusion-based paradigm by employing a shared BERT-type transformer to embed textual and cross-modal features for three reasons: 1) Unlike the visual encoder that usually incorporates pyramid structures (\textit{e}.\textit{g}., Resnet \cite{he2016deep} and PVT \cite{wang2021pyramid}), the text and cross-modal encoder are both BERT-type transformers, whose difference only lies in the number of transformer blocks; 2) The inputs of the text and cross-modal encoder are both token-type embeddings/features ($\mathbf{W}$ and [$\mathbf{T}$, $\mathbf{v}$]). Besides, we hypothesize that the visual features extracted by visual encoders could be treated as high-level semantic tokens as text embeddings; And 3) it is difficult to find a suitable weight initialization for training the cross-modal encoder, which would decrease the pre-training performance as discussed in CLIP4Clip \cite{luo2021clip4clip}. In this way, \textbf{SNP} is as lightweight as twin-tower-based models, and could support various downstream video-text tasks like three-fusion-based ones, achieving a satisfactory balance between pre-training efficiency and performance.

\begin{figure}[t]
	\centering
	\includegraphics[width=0.49\textwidth]{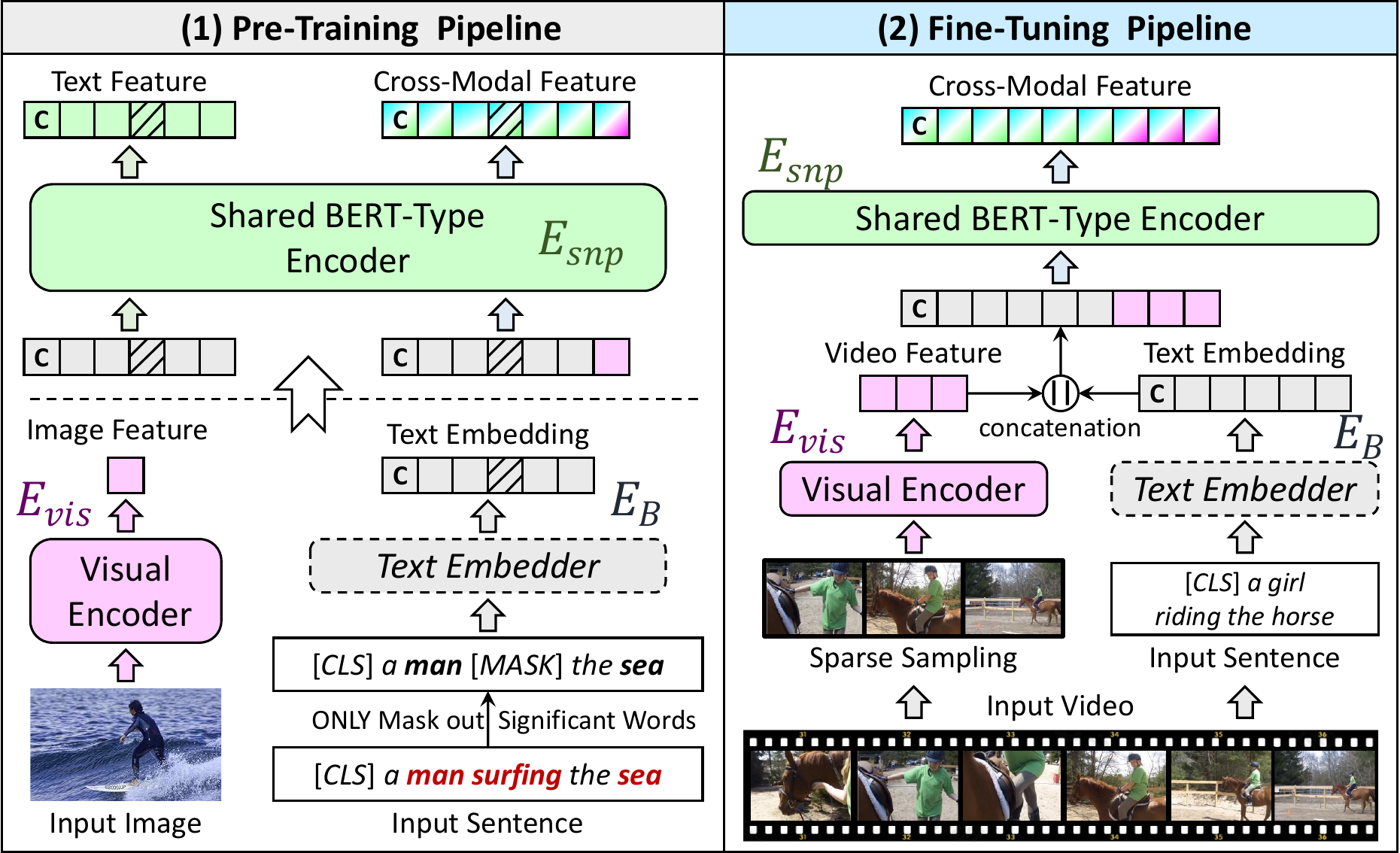}

	\caption{The framework of \textbf{SNP}-$\textbf{S}^\textbf{3}$, which employs a shared BERT-type encoder to process textual and cross-modal features. Following the previous work, we 1) pre-train on image-text datasets, and 2) fine-tune on downstream video-text tasks. We also report the results pre-trained on video-text datasets in Section \ref{sec:performance_compare}. }

	\label{method_pipeline}
\end{figure}

As illustrated in the left part of Figure \ref{method_pipeline}, we denote the shared BERT-type encoder as ${E}_{snp}$, where textual features $\mathbf{T}$ and cross-modal features $\mathbf{M}$ can be calculated as: 

\begin{equation}
\label{SNP-forward}
	\left\{
	\begin{aligned}
		& \mathbf{T} = {E}_{snp}({E}_{B}(S)),
		\\
		& \mathbf{M} = {E}_{snp}([({E}_{B}(S)), {E}_{vis}(I)]).
	\end{aligned}
	\right.
\end{equation}

\subsection{Limitations of Conventional MLM and GVTM}

Proxy tasks directly determine the pre-training objectives, where Masked Language Modeling (MLM) and Global Vision-Text Matching (GVTM) are two widely-employed ones. MLM first masks out a certain percentage of words in a given sentence, and then forces the model to restore these clozes according to visual and textual cues. However, some masking words like prepositions and conjunctions could be easily predicted only by grammar without reviewing the image, which may contribute little to the cross-modal interaction. GVTM aims to learn the cross-modal interaction from image features and global-pooling textual features. However, we believe that global features of a sentence and local information of some inner informative words are equally important, while GVTM only emphasizes the former but omits the latter.

Intuitively, not all words contribute equally to understanding a sentence. Specifically, some words like verbs and nouns are more significant as they provide rich information, while others only act as ``lubricants" to make the pale description ``somebody do something" more fluent and vivid. Based on this intuition, we aim to capture these informative words rather than trivial ones to promote the cross-modal interaction. Towards this end, we propose the Significant Semantic Strengthening ($\textbf{S}^\textbf{3}$) strategy, which includes a novel masking task (MSSM) and a matching one (LVWM) for better pre-training performance.


\subsection{Significant Semantic Mining Algorithm}

\begin{algorithm}[t]
	\caption{Offline Significant Semantic Mining.}
	\label{alg_ossm1}
	\LinesNumbered
	\KwIn{
	BERT vocabulary list ${L}_{BERT}$, \\
	All captions within datasets $\{{Cap}_{i}\}_{i=1}^{N_{data}}$, \\ Pre-defined num ${K}^{ss}$.}
	\KwOut{Significant semantic vocabulary ${L}_{spaCy}$.} 
	
	Set ${L}_{POS}$ = [ 0 ] * ${\rm len}({L}_{BERT})$
	
	\For{$i \leftarrow 1$ \KwTo $N_{data}$}{
	
	    ${T}_{spacy}$ = spaCy.tokenize$({Cap}_{i})$ \;
	    
	    ${P}_{spacy}$ = spaCy.POS$({Cap}_{i})$    \;
	    
	    \For{$j \leftarrow 1$ \KwTo ${\rm len}({T}_{spacy})$}{
	    
	        \If{${T}_{spacy}[\,j\,]$ in ${L}_{BERT}$}{
	        
	            \If{${P}_{spacy}[\,j\,]$ in $[$Verb, Adjective, Noun$]$}{
	            
	                $label$ = BERT.tolabel(${T}_{spacy}[\,j\,]$) \;
	                
	                ${L}_{POS}[\,label\,]$ += 1 \;
		    }
		}
	    }
	}
	
	Set ${\rm Num}_{K}$ = Get-Maximum-K(${L}_{POS}, {K}^{ss}$) \;
	
	Set ${L}_{spaCy}$ = [ 0 ] * ${\rm len}({L}_{BERT})$ \;
	
	\For{$i \leftarrow 1$ \KwTo ${\rm len}({L}_{spaCy})$}{
	
	   \If{${L}_{POS}[\,i\,] \geqslant {\rm Num}_{K}$}{
	        ${L}_{spaCy}[\,i\,]$ = 1   \;
	    }
	}
\end{algorithm}

\begin{algorithm}[t]
	\caption{Online Significant Semantic Mining.}
	\label{alg_ossm2}
	\LinesNumbered
	\KwIn{
	BERT vocabulary list ${L}_{BERT}$, \\
	Significant semantic vocabulary ${L}_{spaCy}$, \\
	Caption $Cap$.}
	\KwOut{Significant semantic chosen list ${L}_{ss}$.}
	Set ${L}_{ss}$ = []    \;
	
    ${T}_{BERT}$ = BERT.tokenize$(Cap)$    \;
    
	\For{$i \leftarrow 1$ \KwTo ${\rm len}({T}_{BERT})$}{
	
	    $label$ = BERT.tolabel(${T}_{BERT}[\,i\,]$) \;
	    
	        \If{${L}_{spaCy}[\,label\,] == 1$}{
                ${L}_{ss} += [\, i \,]$
	    }
	}
\end{algorithm}

We simply define VERBs, NOUNs and ADJECTIVEs as significant semantics since they provide essential information for understanding a sentence. Then the question is how to distinguish these informative words from other trivial ones efficiently. We first attempt to select these significant words by an open-source NLP toolkit spaCy \footnote{Official Website of spaCy: https://spacy.io/} during the pre-training in an end-to-end manner. Unfortunately, it is infeasible for two reasons: 1) The inference time of spaCy is unaffordable, which would slow down the pre-training obviously. And 2) the dictionary of spaCy is quite different from BERT, whose results could not be processed into BERT-type token embeddings directly.

Towards this end, We design a direct and efficient mining algorithm with the open-source NLP toolkit spaCy to find those informative words, and organize them into the significant semantic chosen list ${L}_{ss}$. Specifically, we first maintain an offline significant semantic vocabulary ${L}_{spaCy}$ according to the BERT vocabulary by employing spaCy to review all the captions within pre-training datasets, and then leverage this offline vocabulary ${L}_{spaCy}$ to build the online significant semantic list ${L}_{ss}$ according to the input sentence during the pre-training. The workflow is summarized in Algorithm \ref{alg_ossm1} and Algorithm \ref{alg_ossm2}, where ${\rm len}(\cdot)$ obtains the length of the given list, ${ \rm spaCy.tokenize}(\cdot)$ splits the given caption into tokens, and ${\rm spaCy.POS}(\cdot)$ obtains the parts-of-speech tag of each word by employing spaCy; ${\rm BERT.tolabel}(\cdot)$ translates the given BERT token into the corresponding label number and ${\rm BERT.tokenize}(\cdot)$ tokenizes the given sentence according to the BERT vocabulary; Get-Maximum-K$(\cdot, K)$ aims to find the ${K}^{th}$ maximum number in the given list. We choose the Top 2000 (${K}^{ss}$) vocabs whose parts-of-speech tags are nouns, verbs, or adjectives according to their frequency to build the significant semantic vocabulary ${L}^{spacy}$. We then leverage it to obtain the significant semantic chosen list ${L}_{ss}$.

It takes about half a day to complete the offline significant semantic mining step on all pre-training corpus. Compared with the whole pre-training (usually three days), this time assumption is acceptable. Moreover, the purpose of offline mining step is to build the significant semantic vocabulary, which could be reused after the first building.

\subsection{Overall Optimization Objectives}

\begin{figure}[t]
	\centering
	\includegraphics[width=0.49\textwidth]{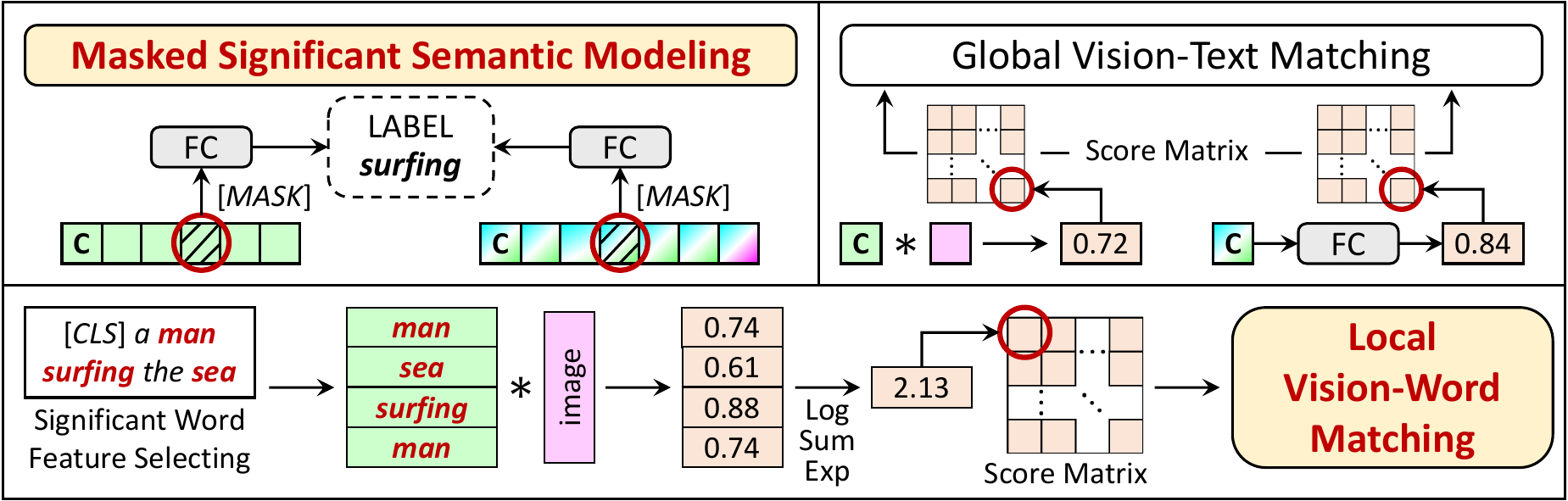}

	\caption{Three types of proxy tasks for pre-training the proposed \textbf{SNP}-$\textbf{S}^\textbf{3}$. Notably, we propose two improved tasks marked in red (MSSM and LVWM) to facilitate the cross-modal interaction. Specifically, MSSM first masks out some significant informative words and forces models to restore these clozes, while LVWM learns the video-text alignment at the word level.}

	\label{method_proxytask}
\end{figure}

Figure \ref{method_proxytask} shows all proxy tasks we employed in our  \textbf{SNP}-$\textbf{S}^\textbf{3}$, including Masked Significant Semantic Modeling (MSSM), Global Vision-Text Matching (GVTM), and Local Vision-Word Matching (LVWM).

\textbf{MSSM} is an improved version of MLM, which only masks out the informative words within the significant semantic chosen list ${L}_{ss}$. Note that all settings (\textit{e}.\textit{g}., the masking rate) except the chosen masked tokens remain the same as the conventional MLM protocol. We calculate MSSM twice for textual features $\mathbf{T}$ ($\mathcal{L}_{1}$) and cross-modal features $\mathbf{M}$ ($\mathcal{L}_{2}$), which can be formulated as follows:

\begin{equation}
\label{loss-mssm-t}
	\mathcal{L}_{1}= \frac{1}{\mathcal{|Q|}} \sum_{q \in \mathcal{Q}} \mathcal{L}_{CE} ({y}^{q}, \mathbf{t}^{q}),
\end{equation}
\begin{equation}
\label{loss-mssm-m}
	\mathcal{L}_{2}= \frac{1}{\mathcal{|Q|}} \sum_{q \in \mathcal{Q}} \mathcal{L}_{CE} ({y}^{q}, \mathbf{m}^{q}),
\end{equation}
where $\mathcal{Q}$ denotes the masked token set, $| \cdot |$ denotes the length of a given set, ${y}^{q}$ denotes the ground-truth token label, and $\mathcal{L}_{CE}$ is the regular Cross-Entropy cost function. 

\textbf{GVTM} aims to model the cross-modal interaction by employing visual features and global-pooling textual features (refer to the hidden state of the first [$cls$] token). Following the conventional paradigm \cite{yang2021taco}, we set two GVTM tasks to align visual and textual features in a parameter-free ($\mathcal{L}_{3}$) and parameter-employed ($\mathcal{L}_{4}$) way:

\begin{equation}
\label{GVTM-non}
	\mathcal{L}_{3}= -\sum_{i=1}^{\mathcal{|B|}} {\rm log}   \frac{ {\rm exp}^{ \langle \mathbf{v}_{i}, \mathbf{t}_{i}^{cls} \rangle } }
	{ {\rm exp}^{ \langle \mathbf{v}_{i}, \mathbf{t}_{i}^{cls} \rangle} + \sum_{j \neq i}{\rm exp}^{ \langle \mathbf{v}_{j}, \mathbf{t}_{i}^{cls} \rangle}   },
\end{equation}
\begin{equation}
\label{GVTM-use}
	\mathcal{L}_{4}= -\sum_{i=1}^{\mathcal{|B|}} {\rm log}   \frac{ {\rm exp}^{ \Theta ( \mathbf{m}_{i, i}^{cls} ) } }
	{ {\rm exp}^{ \Theta ( \mathbf{m}_{i, i}^{cls} )} + \sum_{j \neq i}{\rm exp}^{ \Theta ( \mathbf{m}_{j, i}^{cls} )}   },
\end{equation}
where $\mathcal{|B|}$ is the length of a mini-batch, $\langle \cdot , \cdot \rangle$ denotes the matrix multiplication operation,  $\Theta$ is a Multi-Layer Perception (MLP), and $\mathbf{m}_{j, i}^{cls}$ is the global-pooling cross-modal features of the image-text pair (${I}_{j}, {S}_{i}$). Since $\mathbf{v}, \mathbf{t}^{cls} \in \mathbb{R}^{1*d}$, $\mathcal{L}_{3}$ counts similarity score matrices without introducing any parameters.

\textbf{LVWM} is a complementary to GVTM as it focuses on modeling several informative semantic features at the word level rather than the sentence level. We first build the local significant semantic feature set
$\hat{\textbf{T}} = \{ \hat{\mathbf{t}}^{i} \; | \; \hat{\mathbf{t}}^{i} \in \textbf{T} \}_{i=1}^{{N}_{L}}$ 
from textual features $\textbf{T}$ according to the list ${L}_{ss}$, which contains ${N}_{L}$ significant token features by either random-sampling (if $ {\rm len}({L}_{ss}) > {N}_{L}$) or over-sampling (if ${\rm len}({L}_{ss}) < {N}_{L}$). Then we calculate LVWM loss ($\mathcal{L}_{5}$) as follows:

\begin{equation}
\label{loss-LVWM}
	\mathcal{L}_{5}= -\sum_{i=1}^{\mathcal{|B|}} {\rm log}   \frac{ \sum_{ l=1 }^{ {N}_{L}} {\rm exp}^{ \langle \mathbf{v}_{i}, \hat{\mathbf{t}}_{i}^{l} \rangle } }
	{ \sum_{ l=1 }^{ {N}_{L}} ( {\rm exp}^{ \langle \mathbf{v}_{i}, \hat{\mathbf{t}}_{i}^{l} \rangle} + \sum_{j \neq i}{\rm exp}^{ \langle \mathbf{v}_{j}, \hat{\mathbf{t}}_{i}^{l} \rangle} )   }.
\end{equation}

Ultimately, the objective function of the proposed \textbf{SNP}-$\textbf{S}^\textbf{3}$ is the combination of Eqs. (3)-(7), which is defined as:

\begin{equation}
	\mathcal{L} = \mathcal{L}_{1} + \mathcal{L}_{2} + \mathcal{L}_{3} + \mathcal{L}_{4} + \mathcal{L}_{5}.
\end{equation}

\subsection{Pre-training on Video-Text Datasets} 
\label{pre-train-video-dataset}

We pre-train our \textbf{SNP}-$\textbf{S}^\textbf{3}$ on large video-text datasets to pursue better performance. As a video could be treated as a group of images (frames) in time streams, we sparsely (and randomly) sample ${N}_{V}$ frames from a given video (${N}_{V}$ is usually much smaller than the total number of frames of this video), and set their position embeddings to zero following CLIP4Clip \cite{luo2021clip4clip}. Therefore, the video features $\mathbf{V} = [{\mathbf{v}}^{1}, {\mathbf{v}}^{2}, \cdots, {\mathbf{v}}^{{N}_{V}}]$ could be processed by the visual encoder ${E}_{vis}$. The forward propagation step and optimization objectives (proxy tasks) remain the same as the protocol pre-trained on image-text datasets. \textit{I}.\textit{e}., first replacing image features $\mathbf{v}_{i}$ with video features $\mathbf{V} = \{\mathbf{v}_{i}^{k}\}_{k=1}^{{N}_{V}}$, and then repeating operations of Eqs. (2)-(8).

\subsection{Fine-tuning on Downstream Video-Text Tasks} 

\begin{figure}[t]
	\centering
	\includegraphics[width=0.49\textwidth]{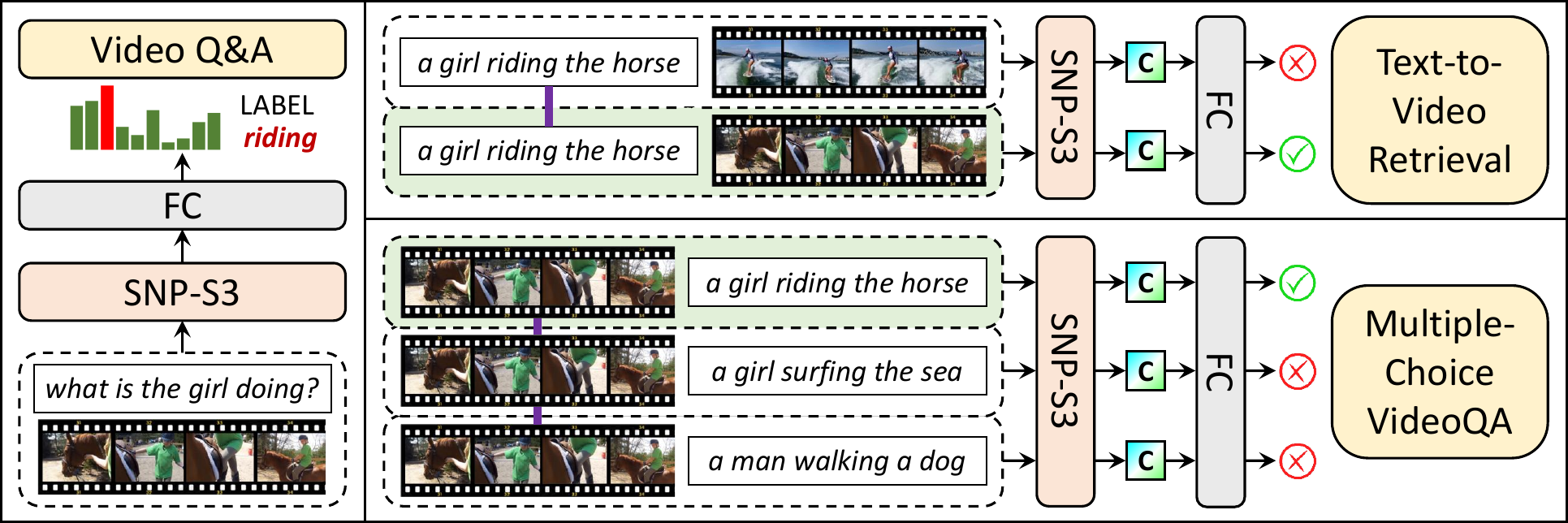}

	\caption{Details of fine-tuning the pre-trained model on three downstream video-text tasks.}

	\label{method_finetune}
\end{figure}

We fine-tune our model on three downstream video-text tasks to evaluate the performance of \textbf{SNP}-$\textbf{S}^\textbf{3}$, the fine-tuning details are illustrated in Figure \ref{method_finetune}. Similar to the pre-training protocol in Section \ref{pre-train-video-dataset}, we randomly sample several frames from raw videos to serve as visual inputs.

\textbf{Text-to-Video Retrieval} (TVR) aims to retrieve the most relevant video according to the input text query. We fine-tune our pre-trained model by reusing two Global Vision-Text Matching objectives $\mathcal{L}_{3}$ (Eq.\ref{GVTM-non}) and $\mathcal{L}_{4}$ (Eq.\ref{GVTM-use}). 

\textbf{Video Question Answering} (VQA) aims to answer natural language questions according to the given videos. In the field of video understanding, VQA is essentially a classification task rather than a generation one. Thus, we add a classifier ${\Theta}_{a}$ on top of the global-pooling cross-modal features $\mathbf{m}^{cls}$ in the last layer of the model. We fine-tune it by calculating the Cross-Entropy loss $\mathcal{L}_{VQA}$ as follows:
\begin{equation}
\label{VQA_loss}
	\mathcal{L}_{VQA} = \mathcal{L}_{CE} ({y}_{a}, {\Theta}_{a}(\mathbf{m}^{cls}) ),
\end{equation}
where ${y}_{a}$ is the label of the ground-truth answer.

\textbf{Multi-Choice Video Question Answering} (MC-VQA) aims to align each video with one out of several candidate answers. Currently, this task is only supported by MSR-VTT multi-choice test set \cite{yu2018joint} without available training data. We directly employ the best model in TVR fine-tuning and traverse this test set once during evaluation, which could be treated as a zero-shot classification task.


\begin{table*}[t]
    \small
    \begin{subtable}[t]{0.33\linewidth}
        \centering
        \begin{tabular}
            {p{2.2cm}<{\centering}|p{2.4cm}<{\centering}}
    		\hline
    		\multicolumn{1}{c|}{Model} &
    		\multicolumn{1}{c}{R@1/5/10 $\uparrow$}
    		\\ \hline
    		 CE & 9.9 / 29.0 / 41.2 \\
    		 CLIPBERT \textit{(p)} & 22.0 / 46.8 / 59.9 \\
    		 HERO$^{*}$ \textit{(f)} & 16.8 / 43.4 / 57.7 \\
    		 CoCoBERT$^{*}$ \textit{(f)} & 22.0 / 48.3 / 61.6 \\
    		 TACo$^{*}$ \textit{(f)} & 24.5 / 52.8 / 65.5 \\
    		 VLM$^{*}$ \textit{(f)} & 28.1 / 55.5 / 67.4    \\
             \hline
            \textbf{SNP-S$^{3}$-PVT} & 26.6 / 55.5 / 67.7   \\
    		\textbf{SNP-S$^{3}$-VST$^{*}$} & \textbf{31.5} / \textbf{61.3} /  \textbf{73.2} \\
    		 \hline
	    \end{tabular}
        \caption{MSRVTT Retrieval (7K-1K split).}
        \label{msrvtt_ret_7k_1k}
    \end{subtable}
    \begin{subtable}[t]{0.33\linewidth}
        \centering
        \begin{tabular}
            {p{2.1cm}<{\centering}|p{2.6cm}<{\centering}}
    		\hline
    		\multicolumn{1}{c|}{Model} &
    		\multicolumn{1}{c}{R@1/5/10 $\uparrow$}
    		\\ \hline
    		 Frozen \textit{(p)} & 25.5 / 54.5 / 66.1 \\
    		 VIOLET \textit{(p)} & 23.5 / 50.5 / 63.9 \\
    		 \textbf{SNP-S$^{3}$-PVT} & \textbf{28.9} / \textbf{57.0} / \textbf{69.4} \\
    		 \hline
    		 MMT & 24.6 / 54.0 / 67.1 \\
    		 TACo$^{*}$ \textit{(f)} & 28.4 / 57.8 / 71.2 \\
    		 Frozen$^{*}$ \textit{(p)} & 32.5 / 61.5 / 71.2 \\
    		 VIOLET$^{*}$ \textit{(p)} & \textbf{34.5} / 63.0 / 73.4  \\
    		 \textbf{SNP-S$^{3}$-VST$^{*}$} & 33.6 / \textbf{65.8} / \textbf{75.1}    \\
    		 \hline
	    \end{tabular}
	\caption{MSRVTT Retrieval (9K-1K split).}
        \label{msrvtt_ret_9k_1k}
    \end{subtable}
    \begin{subtable}[t]{0.33\linewidth}
    \centering
        \begin{tabular}
            {p{2.1cm}<{\centering}|p{2.6cm}<{\centering}}
    		\hline
    		\multicolumn{1}{c|}{Model} &
    		\multicolumn{1}{c}{R@1/5/10 $\uparrow$}
    		\\ \hline
    		 CLIPBERT \textit{(p)} & 20.4 / 48.0 / 60.8 \\
    		 VIOLET \textit{(p)} & 22.8 / 51.2 / 62.0 \\
    		 \textbf{SNP-S$^{3}$-PVT} & \textbf{26.6} / \textbf{57.3} / \textbf{69.1}  \\
    		 \hline
    		 CE & 16.1 / 41.1 / 54.4 \\
    		 MCQ$^{*}$ \textit{(p)} & \textbf{37.0} / 62.2 / 73.9 \\
    		 Frozen$^{*}$ \textit{(p)} & 31.0 / 59.8 / 72.4 \\
    		 VIOLET$^{*}$ \textit{(p)} & 32.6 / 62.8 / 74.7 \\
    		 \textbf{SNP-S$^{3}$-VST$^{*}$} & 34.2 / \textbf{64.2} / \textbf{75.9}    \\
    		 \hline
	    \end{tabular}
        \caption{Didemo Retrieval.}
        \label{didemo}
    \end{subtable}

    \vspace{0.2cm}
    
    \begin{subtable}[t]{0.35\linewidth}
    \centering
        \begin{tabular}
            {p{2.1cm}<{\centering}|p{2.6cm}<{\centering}}
    		\hline
    		\multicolumn{1}{c|}{Model} &
    		\multicolumn{1}{c}{R@1/5/10 $\uparrow$}
    		\\ \hline
    		 CE & 19.8 / 49.0 / 63.8 \\
    		 HERO$^{*}$ \textit{(f)} & 19.2 / 47.4 / 61.8 \\
    		 SSML$^{*}$ \textit{(f)} & 20.3 / 49.0 / 63.3 \\
    		 CoCoBERT$^{*}$ \textit{(f)} & 21.3 / 50.0 / 63.6 \\
    		 Frozen$^{*}$ \textit{(p)} & 33.7 / 64.7 / 76.3 \\
    		 \hline
    		 \textbf{SNP-S$^{3}$-PVT} & 33.1 / 64.5 / 73.7   \\
    		 \textbf{SNP-S$^{3}$-VST$^{*}$} & \textbf{35.1} / \textbf{70.3} / \textbf{80.9} \\
    		 \hline
	    \end{tabular}
        \caption{MSVD Retrieval.}
        \label{msvd-ret}
    \end{subtable}
    \begin{subtable}[t]{0.21\linewidth}
        \centering
        \begin{tabular}
            {p{2.0cm}<{\centering}|p{0.6cm}<{\centering}}
    		\hline
    		\multicolumn{1}{c|}{Model} &
    		\multicolumn{1}{c}{Acc}
    		\\ \hline
    		 HCRN & 35.6 \\
             CLIPBERT \textit{(p)} & 37.4 \\
    		 SSML$^{*}$ \textit{(f)} & 35.1 \\
    		 JustAsk$^{*}$ \textit{(f)} & 41.5 \\
    		 ALPRO$^{*}$ \textit{(p)} & 42.1 \\
    		 \hline
    		 \textbf{SNP-S$^{3}$-PVT} & 42.0 \\
    		 \textbf{SNP-S$^{3}$-VST$^{*}$} & \textbf{43.1} \\
    		 \hline
	    \end{tabular}
        \caption{MSRVTT-QA.}
        \label{msrvtt_qa}
    \end{subtable}
    \begin{subtable}[t]{0.21\linewidth}
        \centering
        \begin{tabular}
            {p{2.0cm}<{\centering}|p{0.6cm}<{\centering}}
    		\hline
    		\multicolumn{1}{c|}{Model} &
    		\multicolumn{1}{c}{Acc}
    		\\ \hline
    		 DualVGR & 39.0 \\
    		 SSML$^{*}$ \textit{(f)} & 35.1 \\
    		 CoMVT$^{*}$ \textit{(f)} & 42.6 \\
    		 JustAsk$^{*}$ \textit{(f)} & 46.3 \\
    		 ALPRO$^{*}$ \textit{(p)} & 45.9 \\
    		 \hline
    		 \textbf{SNP-S$^{3}$-PVT} & 46.2 \\
    		 \textbf{SNP-S$^{3}$-VST$^{*}$} & \textbf{47.1} \\
    		 \hline
	    \end{tabular}
        \caption{MSVD-QA.}
        \label{msrvtt_qa}
    \end{subtable}
    \begin{subtable}[t]{0.21\linewidth}
        \centering
        \begin{tabular}
            {p{2.0cm}<{\centering}|p{0.6cm}<{\centering}}
    		\hline
    		\multicolumn{1}{c|}{Model} &
    		\multicolumn{1}{c}{Acc}
    		\\ \hline
    		 JSFusion & 83.4 \\
    		 CLIPBERT \textit{(p)} & 88.2 \\
    		 MERLOT$^{*}$ \textit{(f)} & 90.9 \\
    		 VLM$^{*}$ \textit{(f)} & 91.6 \\
    		 VIOLET$^{*}$ \textit{(p)} & 91.9 \\
    		 \hline
    		 \textbf{SNP-S$^{3}$-PVT} & \textbf{92.3}   \\
    		 \textbf{SNP-S$^{3}$-VST$^{*}$} & \textbf{96.5} \\
              \hline
	    \end{tabular}
        \caption{MSRVTT MC-VQA Set.}
        \label{msrvtt-mcvqa}
    \end{subtable}

    \caption{Performance comparison of different methods on three downstream video-text tasks and six corresponding datasets. The superscript ``*'' denotes that the method is pre-trained on large-scale video datasets (\textit{e}.\textit{g}., WebVid). \textit{p} and \textit{f} denote the methods belong to pixel-level pre-training and feature-level ones. The suffix ``PVT'' and ``VST'' represent the model whose visual encoder is based upon PVTv2-B2 (pre-trained on COCO+VG) and VideoSwin-B (pre-trained on CC+WebVid). Note that some methods only conduct experiments on a certain range of datasets. Thus, Baselines on different datasets may vary a lot.}
\label{result_all}
\vspace{0.2cm}
\end{table*}

\section{Experiments}
\label{sec:experiment}

\subsection{Datasets}

\subsubsection{Pre-training Datasets}

We pre-train our \textbf{SNP}-$\textbf{S}^\textbf{3}$ on three large-scale image-text datasets: \textbf{COCO} \cite{lin2014microsoft}, \textbf{Visual Genome} \cite{krishna2017visual} (VG), and \textbf{Conceptual Captions} \cite{sharma2018conceptual} (CC), which contain more than 0.59M, 5.4M, and 3.1M image-text pairs, respectively.

Besides these image-text datasets, we also pre-train our \textbf{SNP}-$\textbf{S}^\textbf{3}$ on WebVid \cite{bain2021frozen}, a large-scale video-text dataset including 2.5M video-text pairs, to pursue better fine-tuning performance on three downstream tasks.

\subsubsection{Fine-tuning Datasets}

We fine-tune our pre-trained model on three downstream video-text tasks, including six corresponding datasets.

\textbf{Text-to-Video Retrieval} : 1) \textit{MSR-VTT} \cite{xu2016msr} contains 10K video clips associated with 200K sentences. There are two widely-employed validation splits, one takes 7K videos for training and randomly selects 1K videos from the remaining ones for testing (7K-1K split), while the other uses 9K videos for training and the remaining 1K for testing (9K-1K split). In this paper, we report the fine-tuning results on these two splits. 2) \textit{DiDeMo} \cite{anne2017localizing} contains 10K Flickr videos associated with 40K sentences. 3) \textit{MSVD} \cite{chen-dolan-2011-collecting} contains 2K video clips associated with 80K descriptions.

\textbf{Video-Question Answering}: 4) \textit{MSRVTT-QA} \cite{xu2017video} is built upon MSR-VTT and contains 10K videos with 243K open-ended questions and 1.5K answer classes. 5) \textit{MSVD-QA} \cite{xu2017video} is built upon MSVD and contains 2K videos with 50K open-ended questions and 2.4K answer classes. 

\textbf{Multi-Choice Video Question Answering}: 6) \textit{MSR-VTT Multi-Choice Test Set} \cite{yu2018joint} contains 3K videos. Each video has five candidates with one correct answer.

\textbf{Metrics}: For TVR, we employ Recall@K (R@K) and Median Rank (MdR) to measure the text-to-video retrieval performance. For VQA and MC-VQA, we employ Accuracy (Acc) to evaluate the answering correctness.

\begin{table*}
	\small
	\centering
	\begin{tabular}{p{0.6cm}<{\centering}|p{1.5cm}<{\centering}|p{2.5cm}<{\centering}|p{1.0cm}<{\centering}
	|p{2.8cm}<{\centering}p{1.5cm}<{\centering}
	|p{2.8cm}<{\centering}p{1.5cm}<{\centering}}
		\hline
		\multicolumn{1}{c|}{\multirow{2}{*}{No.}} &
		\multicolumn{1}{c|}{Model} &
		\multicolumn{1}{c|}{Parameters} &
		\multicolumn{1}{c|}{\multirow{2}{*}{Losses}} &
		\multicolumn{2}{c|}{MSRVTT (7K-1K split)} &
		\multicolumn{2}{c}{MSVD}
		\\ \cline{5-8} 
		\multicolumn{1}{c|}{} &
		\multicolumn{1}{c|}{Name} &
		\multicolumn{1}{c|}{Count} &
		\multicolumn{1}{c|}{} &
		\multicolumn{1}{c}{R@1/5/10 $\uparrow$ (MdR $\downarrow$)} &
		\multicolumn{1}{c|}{QA: Acc} &
		\multicolumn{1}{c}{R@1/5/10 $\uparrow$ (MdR $\downarrow$)} &
		\multicolumn{1}{c}{QA: Acc}
		\\ \hline
		
		A1 & P3E-R50 & 205.5M & & 18.3 / 46.0 / 58.8 (7) &40.40 & 24.5 / 49.8 / 64.2 (6) & 40.99  \\
		A2 & SNP-R50 & 160.4M (\textbf{-22.0\%}) & MLM, & 21.7 / 47.8 / 61.4 (6) & 40.96 & 22.8 / 53.9 / 66.1 (5) & 43.63
		\\ \cline{2-3} \cline{5-8}
		A3 & P3E-PVT & 206.2M & GVTM & 22.7 / 48.0 / 62.5 (6) & 40.76 & 26.4 / 54.6 / 67.8 (4) & 43.83 \\
		A4 & SNP-PVT & 161.1M (\textbf{-21.9\%}) &  & 25.0 / 52.3 / 64.1 (5) & 41.44 & 28.2 / 58.7 / 70.8 (3) & 44.71 \\
		
		 \hline
		
	\end{tabular}
    \caption{Ablation study of the conventional three-fusion-based pixel-level pre-training paradigm (P3E) and our proposed version (SNP) that includes a shared BERT-type encoder on two downstream video-text tasks (TVR and VQA) of two datasets (MSRVTT and MSVD). ``R50'' and ``PVT'' represent that the model utilizes Resnet-50 and PVTv2-B2 as the basic visual encoder.}
\label{ablation_ant}
\end{table*}

\begin{figure*}
	\centering
	\begin{subfigure}{0.245\linewidth}
		\includegraphics[width=1.0\textwidth]{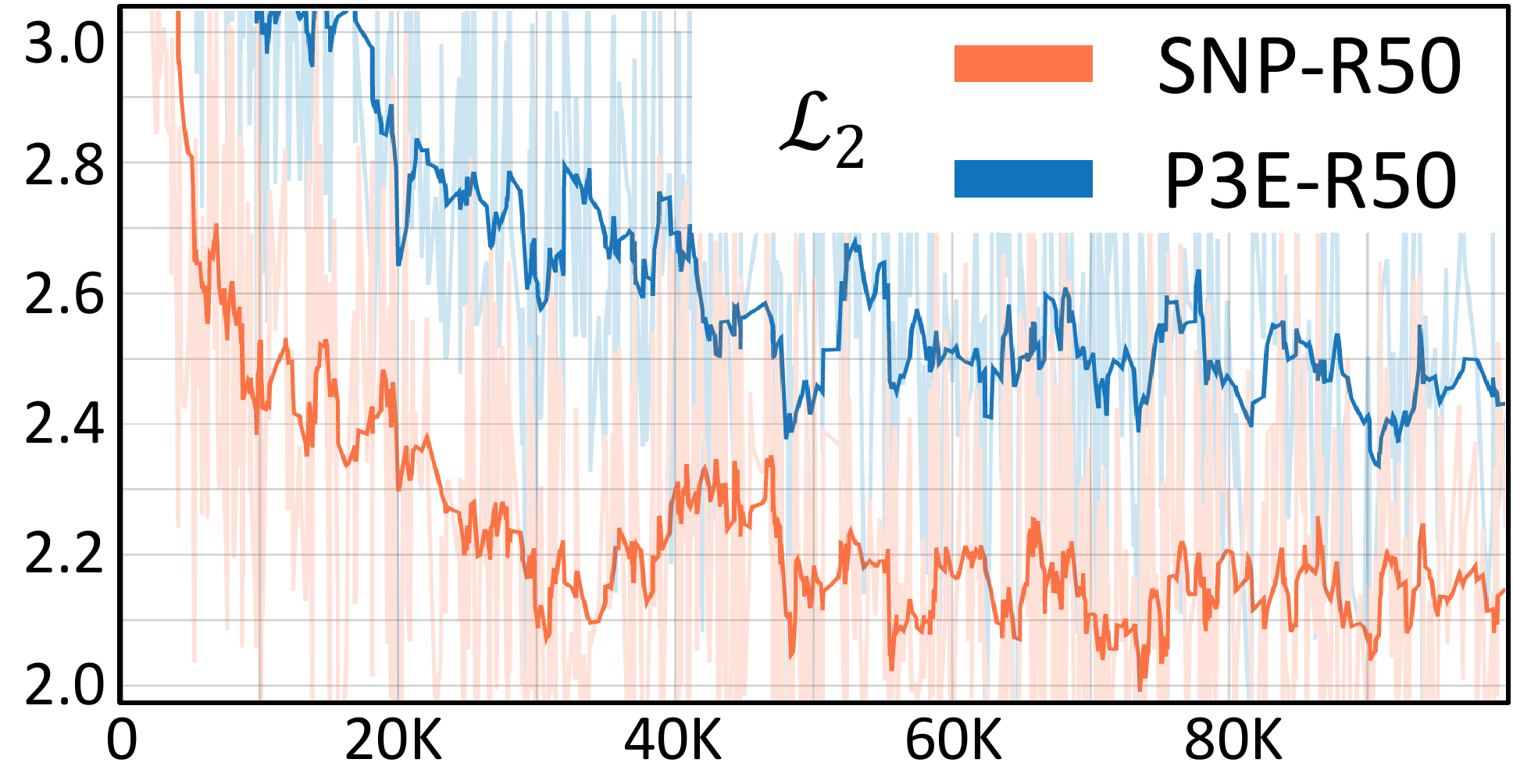}
		\caption{Masking loss value of $\mathcal{L}_{2}$.}
	\end{subfigure}
 	\begin{subfigure}{0.245\linewidth}
		\includegraphics[width=1.0\textwidth]{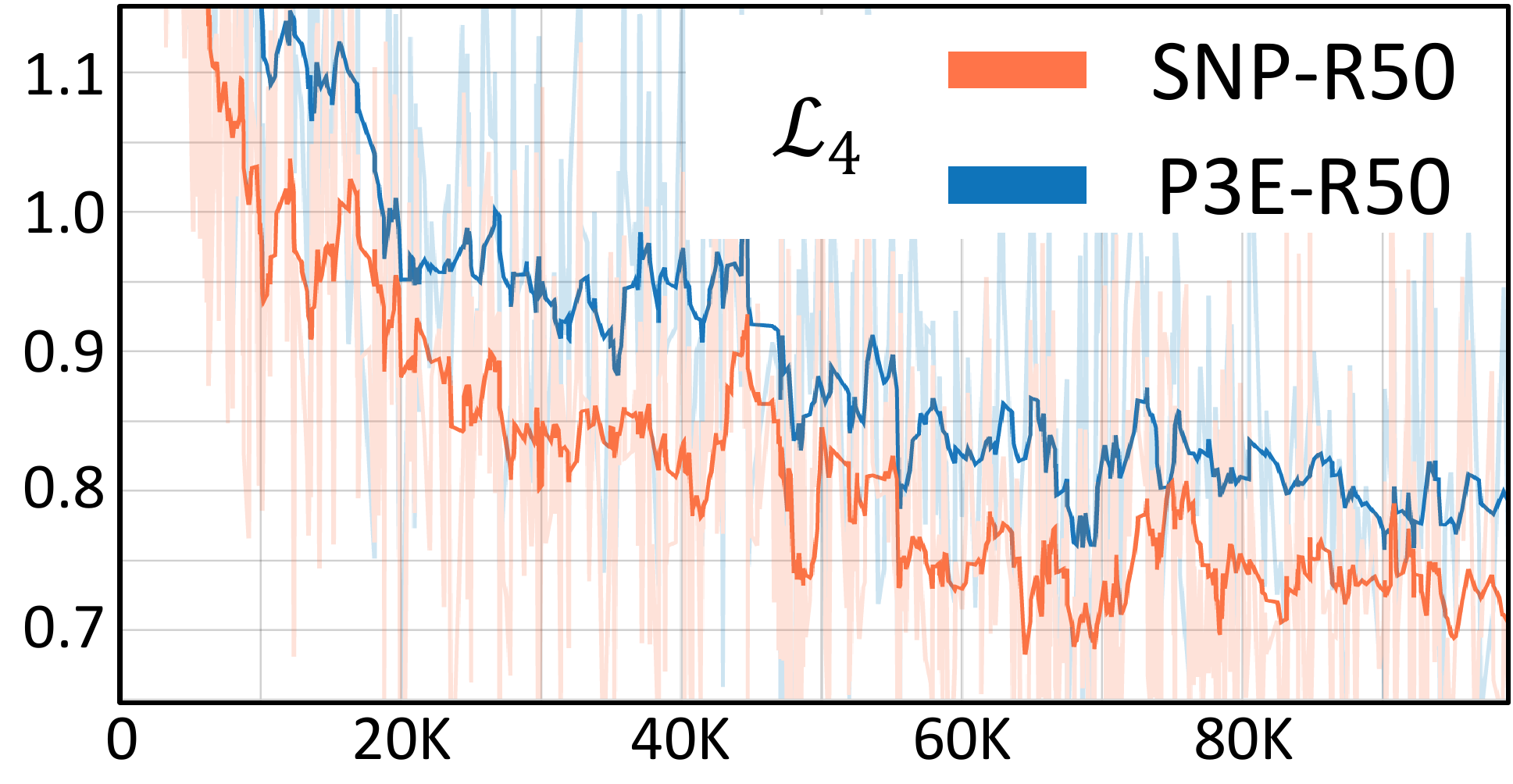}
		\caption{Matching loss value of $\mathcal{L}_{4}$.}
	\end{subfigure}
 	\begin{subfigure}{0.245\linewidth}
		\includegraphics[width=1.0\textwidth]{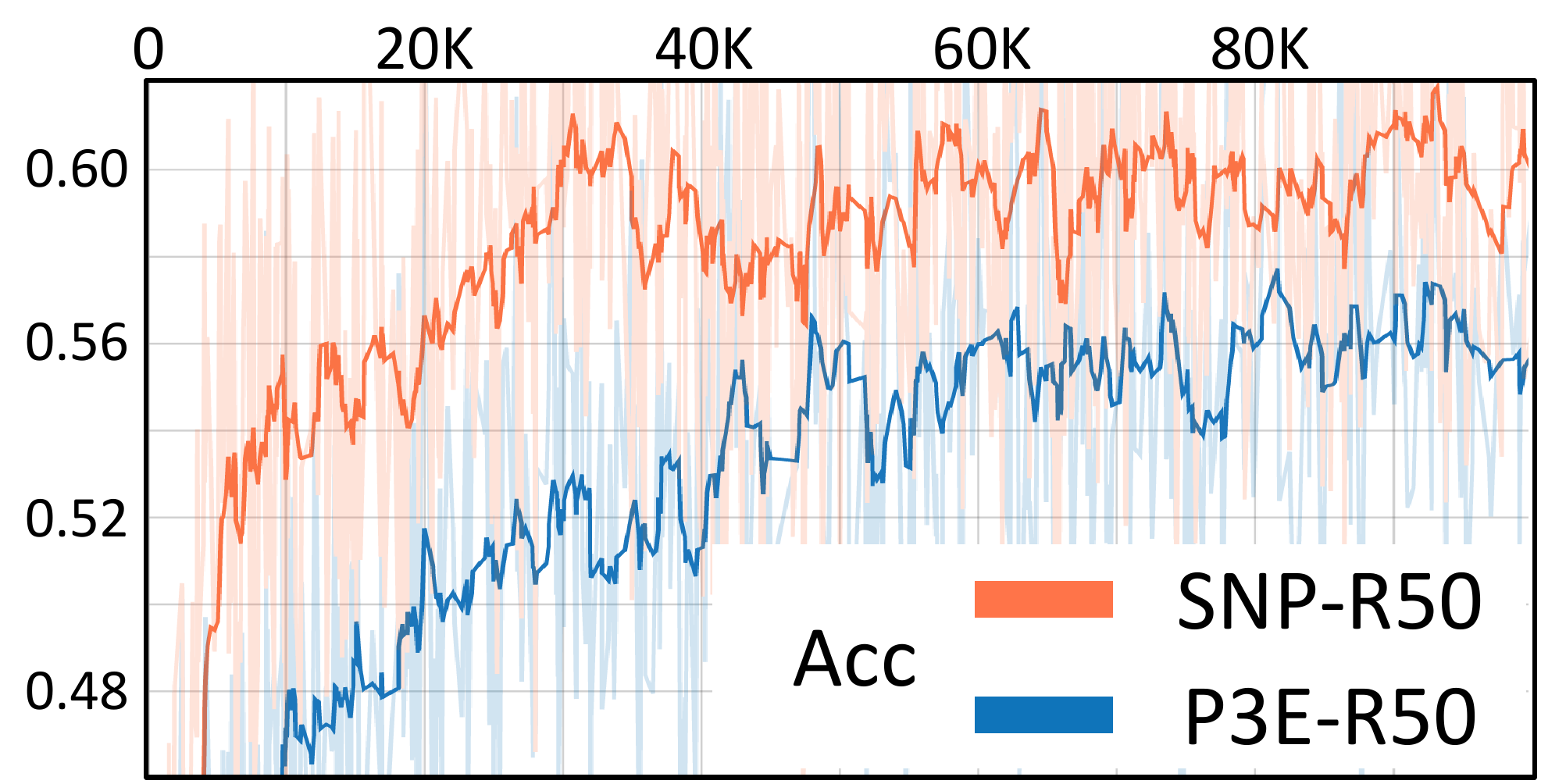}
		\caption{Masking metric of $\mathcal{L}_{2}$: Acc.}
	\end{subfigure}
	\begin{subfigure}{0.245\linewidth}
		\includegraphics[width=1.0\textwidth]{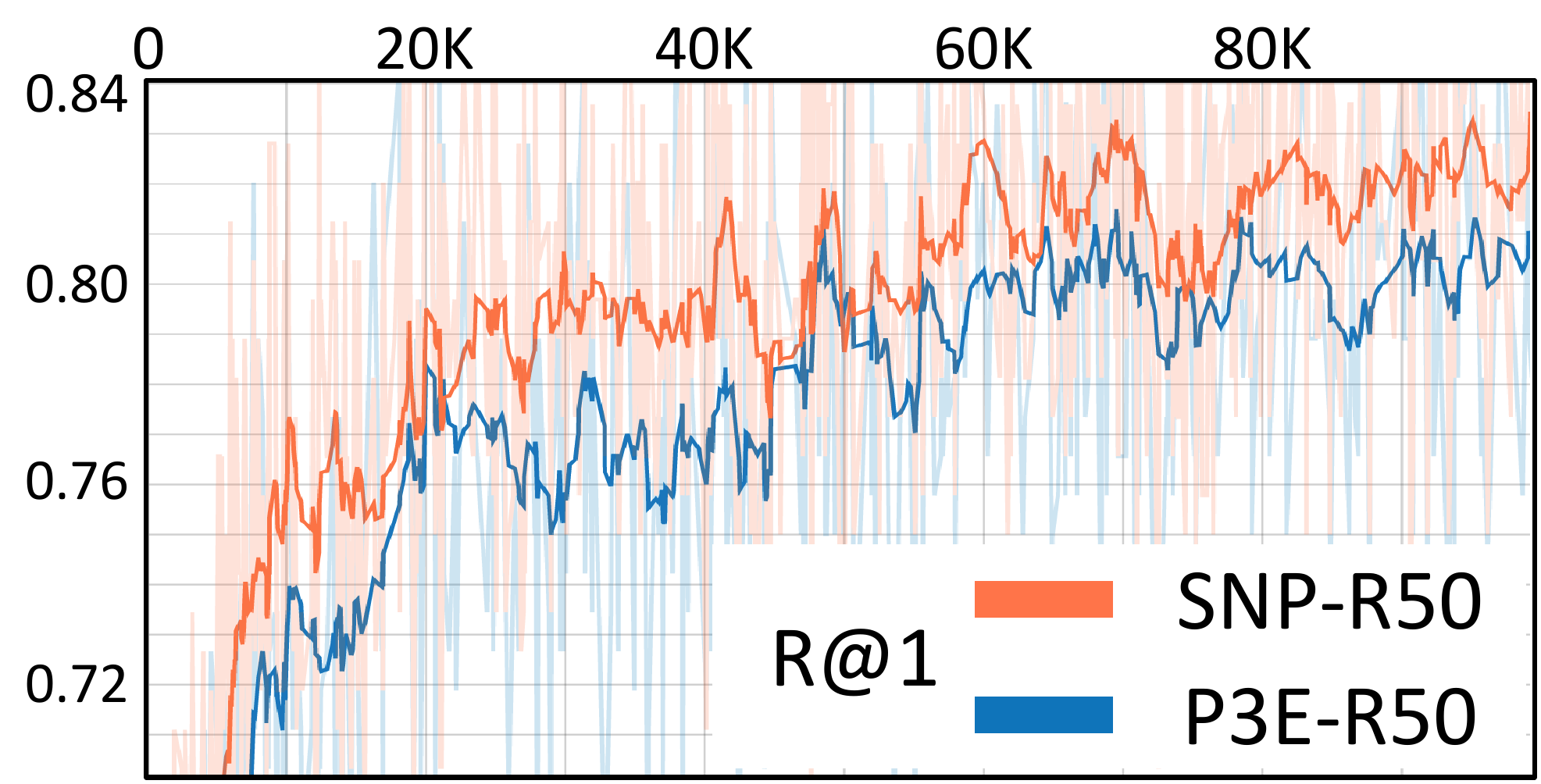}
		\caption{Matching metric of $\mathcal{L}_{4}$: R@1.}
	\end{subfigure}
	
	\vspace{0.2cm}

 	\begin{subfigure}{0.245\linewidth}
		\includegraphics[width=1.0\textwidth]{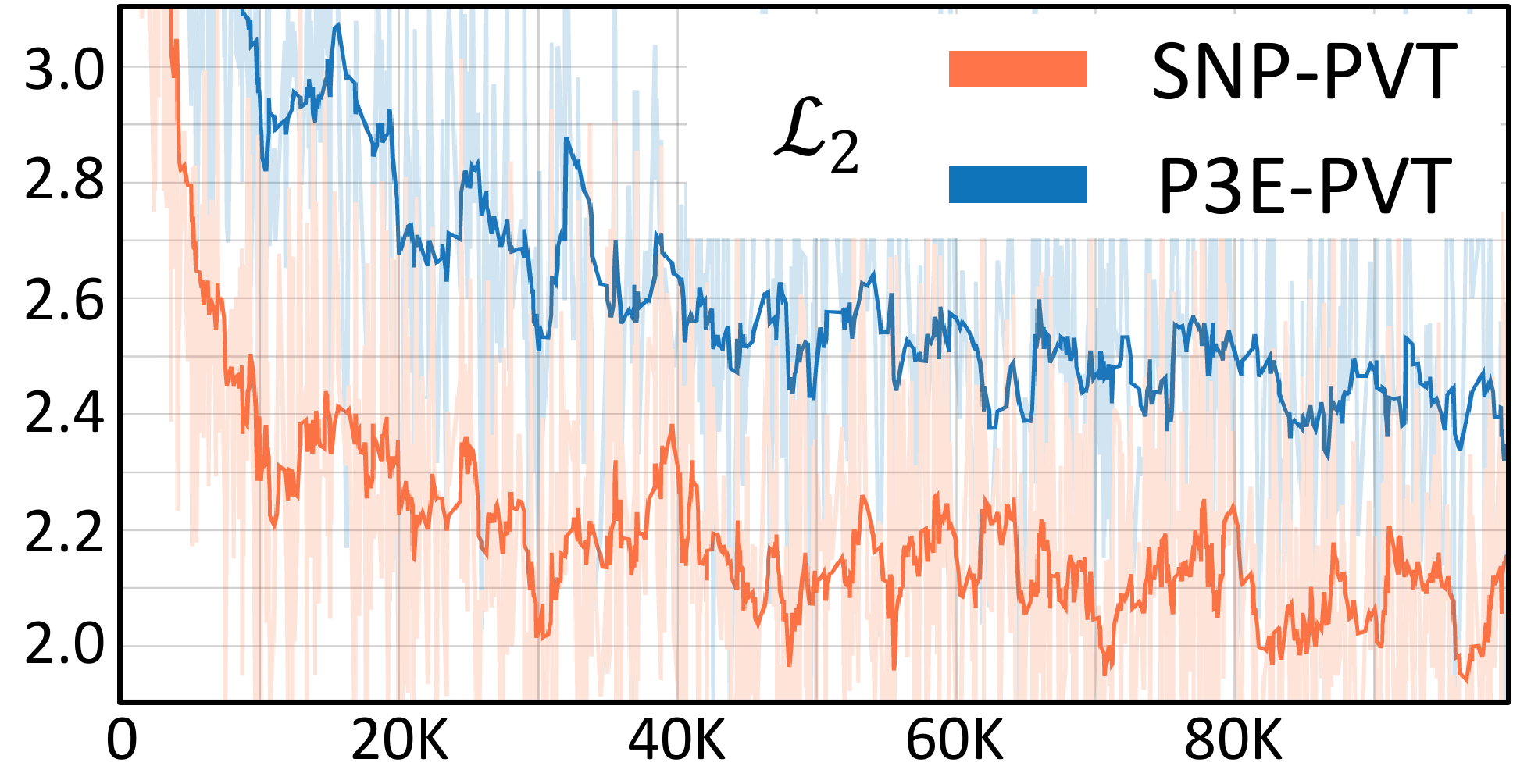}
		\caption{(e) Masking loss value of $\mathcal{L}_{2}$.}
	\end{subfigure}
	\begin{subfigure}{0.245\linewidth}
		\includegraphics[width=1.0\textwidth]{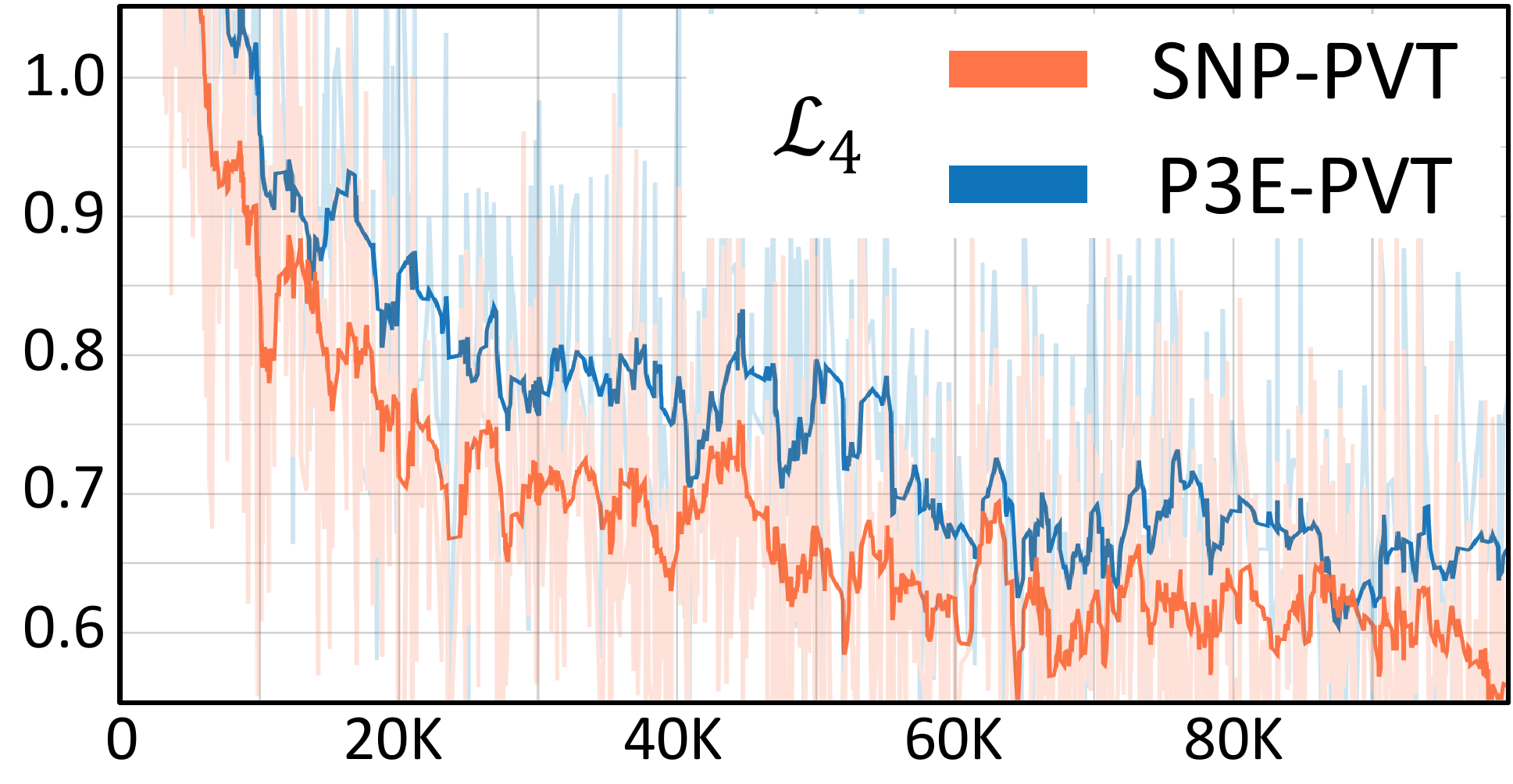}
		\caption{(f) Matching loss value of $\mathcal{L}_{4}$.}
	\end{subfigure}
 	\begin{subfigure}{0.245\linewidth}
		\includegraphics[width=1.0\textwidth]{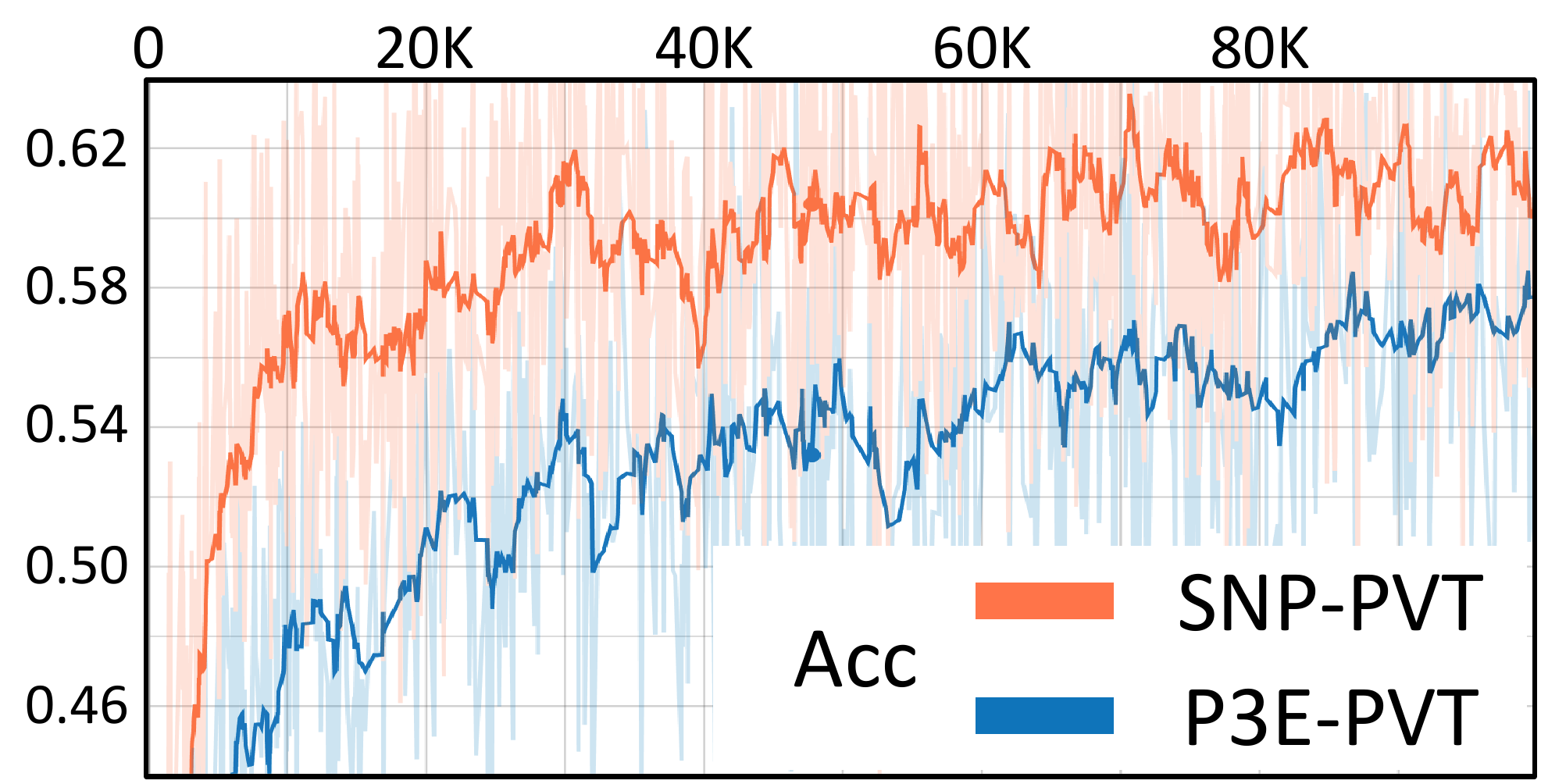}
		\caption{(g) Masking metric of $\mathcal{L}_{2}$: Acc.}
	\end{subfigure}
	\begin{subfigure}{0.245\linewidth}
		\includegraphics[width=1.0\textwidth]{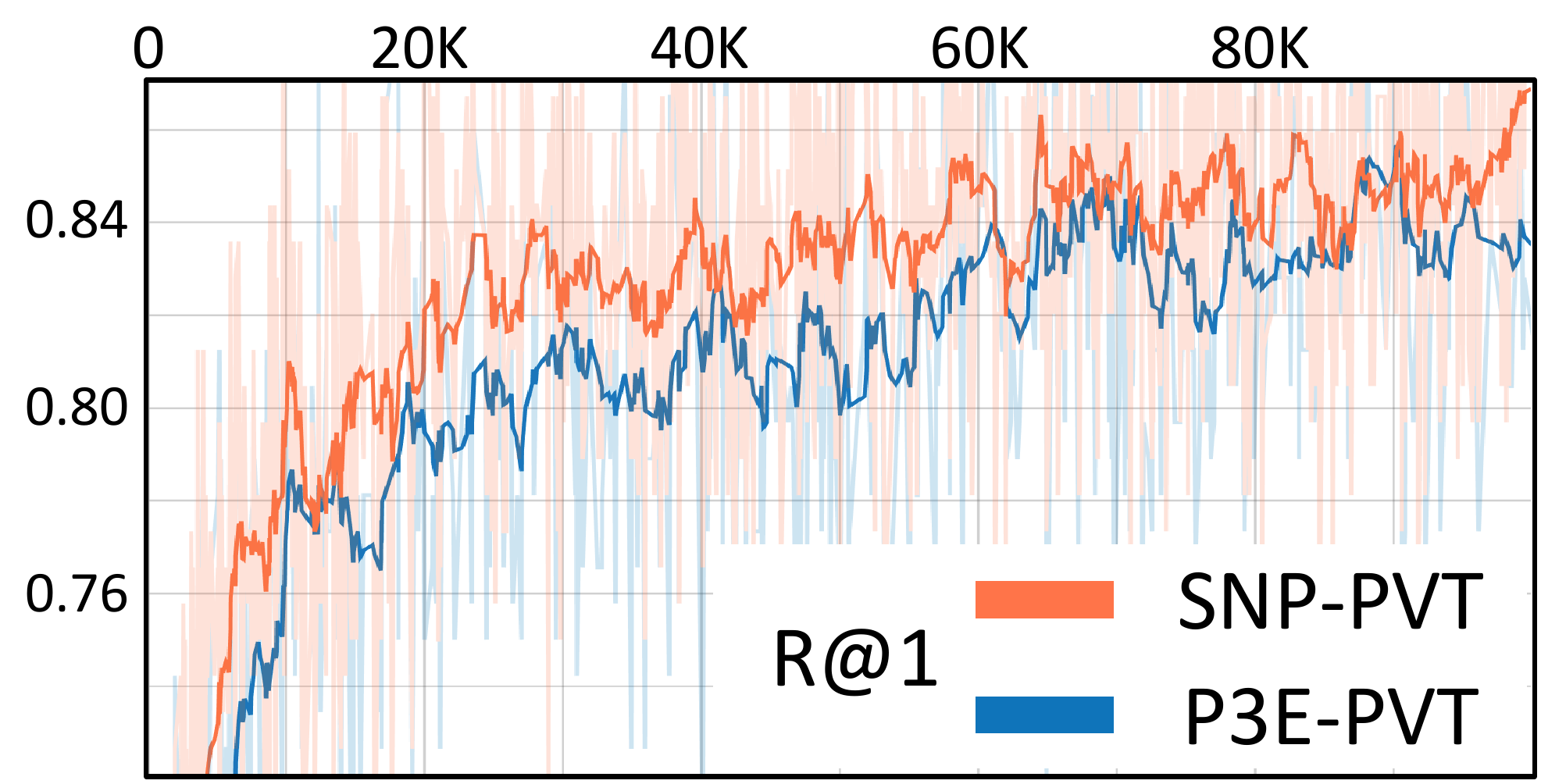}
		\caption{(h) Matching metric of $\mathcal{L}_{4}$: R@1.}
	\end{subfigure}
	
	\caption{Visualization results of losses and evaluation performance towards masking ($\mathcal{L}_{2}$) and matching ($\mathcal{L}_{4}$) objectives in TensorBoard. The smoothing rate is 0.9, and we visualize the first 100K steps of the pre-training.}
	
	\vspace{0.1cm}
	\label{ablation_tensorboard}
\end{figure*}

\begin{table*}[t]
	\small
	\centering
        
	\begin{tabular}{p{0.6cm}<{\centering}|p{1.0cm}<{\centering}|p{1.5cm}<{\centering}p{2.5cm}<{\centering}
	|p{2.8cm}<{\centering}p{1.5cm}<{\centering}
	|p{2.8cm}<{\centering}p{1.5cm}<{\centering}}
		\hline
		\multicolumn{1}{c|}{\multirow{2}{*}{No.}} &
		\multicolumn{1}{c|}{Model} &
		\multicolumn{1}{c}{Masking} &
		\multicolumn{1}{c|}{Matching} &
		\multicolumn{2}{c|}{MSRVTT (7K-1K split)} &
		\multicolumn{2}{c}{MSVD}
		\\ \cline{5-8} 
		\multicolumn{1}{c|}{} &
		\multicolumn{1}{c|}{Name} &
		\multicolumn{1}{c}{Loss} &
		\multicolumn{1}{c|}{Loss(es)} &
		\multicolumn{1}{c}{R@1/5/10 (MdR)} &
		\multicolumn{1}{c|}{QA: Acc} &
		\multicolumn{1}{c}{R@1/5/10 (MdR)} &
		\multicolumn{1}{c}{QA: Acc}
		\\ \hline
		
		B1 & & MLM & GVTM & 21.7 / 47.8 / 61.4 (6)  &40.96 & 22.8 / 53.9 / 66.1 (5) & 43.63 \\
		B2 & SNP & MSSM & GVTM & 20.9 / 49.3 / 63.3 (6) & 41.21 & 24.6 / 55.4 / 67.5 (4) & 44.42 \\
		B3 & R50 & MLM & GVTM, LVWM & 21.2 / 49.3 / 62.8 (6) &41.22 & 24.9 / 55.8 / 69.1 (4) & 44.22 \\
		B4 & & MSSM & GVTM, LVWM & \textbf{23.7} / \textbf{51.2} / \textbf{64.6}  (\textbf{5}) &\textbf{41.47} & \textbf{27.3} / \textbf{56.9} / \textbf{70.2} (\textbf{4}) & \textbf{44.87} \\
		\hline
		B5 & & MLM & GVTM & 25.0 / 52.3 / 64.1 (5) & 41.44 & 28.2 / 58.7 / 70.8 (3) & 44.71 \\
		B6 & SNP & MSSM & GVTM & \textbf{27.2} / 53.2 / 65.7 (5) &41.74 & 31.0 / 59.0 / 71.8 (3) & 45.69\\
		B7 & PVT & MLM & GVTM, LVWM & 25.5 / 52.4 / 64.7 (5) & 41.72 & 30.3 / 60.5 / 71.2 (3) & 45.31\\
		B8 & & MSSM & GVTM, LVWM & 26.6 / \textbf{55.5} / \textbf{67.7}  (\textbf{4}) &\textbf{42.00} & \textbf{33.1} / \textbf{64.5} / \textbf{73.7} (\textbf{3}) & \textbf{46.18}\\
		\hline
		B9 & SNP & MLM & GVTM &  28.7 / 59.4 / 71.1 (4) & 42.52 & 33.4 / 66.7 / 78.5 (3) & 45.42 \\
		B10 & VST & MSSM & GVTM, LVWM & \textbf{31.5} / \textbf{61.3} / \textbf{73.2}  (\textbf{3}) &\textbf{43.09} & \textbf{35.1} / \textbf{70.3} / \textbf{80.9} (\textbf{2}) & \textbf{47.15}\\
		
		 \hline
	\end{tabular}
    \caption{Ablation study of employing different combinations of masking and matching losses. The proposed ${\rm \textbf{S}}^{\textbf{3}}$ strategy includes a novel masking loss (Masked Significant Semantic Modeling, MSSM) and a matching one (Local Vision-Word Matching, LVWM).}
\vspace{-0.2cm}
\label{ablation_sss}
\end{table*}

\subsection{Experimental Settings}

\subsubsection{Pre-training Implementation Details}

We employ three types of visual encoders, namely Resnet-50 (R50) \cite{he2016deep}, Pyramid Vision Transformer (PVT) \cite{wang2021pyramid}, and Video Swin Transformer (VST) \cite{liu2022video}. We initialize these three modules with the parameters pre-trained on ImageNet \cite{deng2009imagenet}. For the shared BERT-type network that processes textual and cross-modal features, we utilize the BERT-Base version with 12 layers of transformer blocks. We initialize this BERT-type module with the parameters pre-trained on BookCorpus \cite{zhu2015aligning} and English Wikipedia. Note that \textbf{SNP-R50} and \textbf{SNP-PVT} are pre-trained on COCO+VG (5.6M image-text pairs), while \textbf{SNP-VST} is pre-trained on CC+WebVid (5.5M image/video-text pairs).

For hyper-parameters, we set the length of text tokens ${N}_{t} = 30$, and the dimension of the hidden state $d = 768$. For the proposed MSSM task, the masking rate is 15$\%$. For the proposed LVWM task, we set the number of the chosen significant tokens ${N}_{L} = 3$. We sparsely sample 4 frames from raw videos when pre-training on video-text datasets.

We pre-train our \textbf{SNP}-$\textbf{S}^\textbf{3}$ by the Adam optimizer with a momentum of 0.9. The total pre-training stage lasts for 200,000 steps with a batch size of 128. The initial learning rate is 5e-5 and is decayed by a factor of 10 after 110,000 iterations. The whole pre-training takes about 3 days to complete on 8 NVIDIA V100 GPUs.

\subsubsection{Fine-tuning Implementation Details}

For all three downstream video-text tasks, the optimizer and hyper-parameters remain the same as the pre-training configuration. Following \cite{lei2021less}, we sparsely sample 16 frames from input videos for fine-tuning and testing. The total fine-tuning stage lasts for 15,000 steps. The batch size is set to 32. The initial learning rate is set to 1e-5. 

\subsection{Performance Comparison}
\label{sec:performance_compare}

We compare our \textbf{SNP}-$\textbf{S}^\textbf{3}$ with various state-of-the-art baselines, including the following methods: 

\begin{itemize}

	\item Methods without pre-training: MMT \cite{huang2020multimodal}, CE \cite{liu2019use}, HCRN \cite{le2020hierarchical}, DualVGR \cite{wang2021dualvgr}, and JSFusion \cite{yu2018joint}.
 
	\item Pixel-level pre-training methods: CLIPBERT (5.6M) \cite{lei2021less}, Frozen (5.5M) \cite{bain2021frozen}, VIOLET (185M) \cite{fu2021violet}, MCQ (5.5M) \cite{ge2022bridging}, and ALPRO (5.5M) \cite{li2022align}. The number in ``()" denotes the volume of video/image-text pairs within the pre-training corpus. We use 5.6M image-text pairs in our Resnet/PVT version and 5.5M image/video-text pairs in our VST version for a fair comparison.
 
	\item Feature-level pre-training methods: HERO \cite{li2020hero}, VLM \cite{xu2021vlm}, TACo \cite{yang2021taco}, SSML \cite{amrani2021noise}, CoMVT \cite{seo2021look}, JustAsk \cite{yang2021just}, MERLOT \cite{zellers2021merlot}, and CoCoBERT \cite{luo2021coco}.

\end{itemize}

Table \ref{result_all} presents detailed experimental results on three downstream video-text tasks (TVR, VQA, and MC-VQA) and six corresponding datasets. Note that some methods only conduct experiments on a certain range of datasets (\textit{e}.\textit{g}., Frozen reports its results on MSRVTT, Didemo, and MSVD in its paper). Thus, baselines on different datasets may vary a lot. We report the results of our methods pre-trained on image-text datasets (\textbf{SNP-S$^{3}$-PVT}) and video-text datasets (\textbf{SNP-S$^{3}$-VST$^{*}$}). We have several observations as follows:

\begin{itemize}

	\item \textbf{SNP-S$^{3}$-VST$^{*}$} achieves the best performance among all pixel-level and feature-level video pre-training methods. We outperform Frozen \cite{lei2021less} by 3.9$\%$, 3.5$\%$, and 4.6$\%$ at R@10 on the TVR tasks of MSRVTT (9K-1K), Didemo, and MSVD, respectively.
 
	\item \textbf{SNP-S$^{3}$-PVT} outperforms other pixel-level pre-training methods pre-trained on image-text datasets by a large margin. On the MSRVTT dataset, we outperform current SOTA baselines by 7.8$\%$ (CLIPBERT) on R@10 of the 7K-1K split, 3.3$\%$ (Frozen) on R@10 of the 9K-1K split, 4.6$\%$ (CLIPBERT) on the Accuracy of VQA, and 4.1$\%$ (CLIPBERT) on the Accuracy of MC-VQA. 
 
	\item \textbf{SNP-S$^{3}$-PVT} that pre-trained on image-text datasets shows comparable performance with other methods pre-trained on video-text datasets. \textit{E}.\textit{g}., \textbf{SNP-S$^{3}$-PVT} outperforms TACo at all metrics on the MSRVTT 7K-1K split by more than 2.0$\%$. 

\end{itemize}

\begin{figure*}[t]
	\centering
	\includegraphics[width=1.0\textwidth]{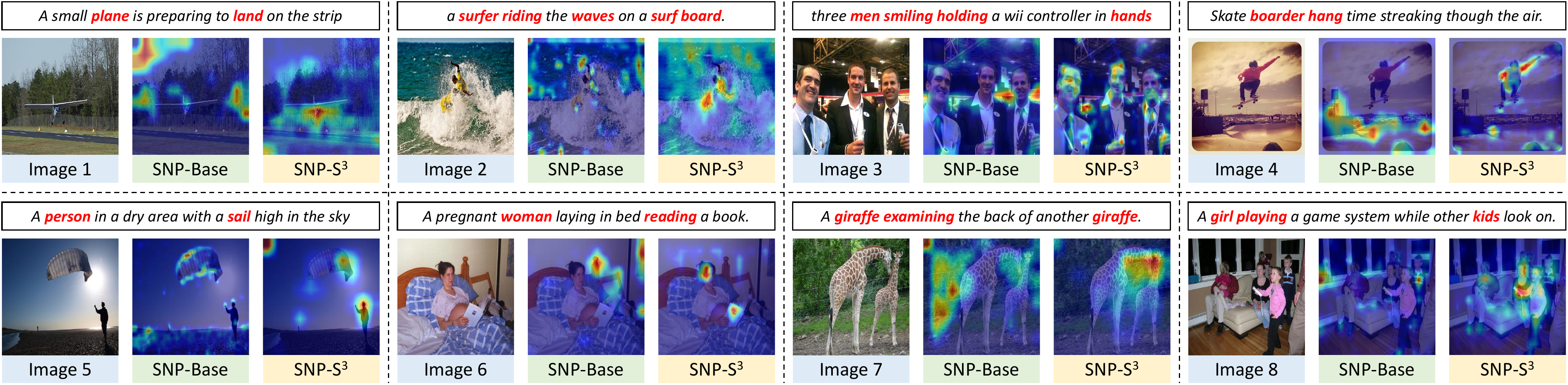}

	\caption{Qualitative analysis of the proposed Significant Semantic Strengthen (\textbf{S$^{3}$}) strategy between SNP-PVT and SNP-S$^{3}$-PVT. We visualize the attention localization map of the last convolution layer in PVT by the toolkit Grad-CAM. We use red to mark the informative words emphasized by SNP-S$^{3}$-PVT while omitted by SNP-PVT.}
	\label{visual_grad}

 \vspace{-0.2cm}

\end{figure*}


\subsection{Ablation Study of SNP}

As aforementioned, the proposed Shared Network Pre-training (\textbf{SNP}) combines the advantages of two mainstream pixel-level architectures, which is lightweight and could support various downstream video-text tasks.

Table \ref{ablation_ant} compares the fine-tuning performance on several downstream video-text tasks between the conventional three-fusion-based pixel-level paradigm (P3E) and our proposed version (SNP), while Figure \ref{ablation_tensorboard} compares losses and evaluation performance during pre-training. Since the only difference between SNP and P3E is how to build its cross-modal encoder, where SNP shares the same parameters with the text encoder while P3E utilizes a separate one, so we only compare losses and metrics ($\mathcal{L}_{2}$ in Eq.\ref{loss-mssm-m} and $\mathcal{L}_{4}$ in Eq.\ref{GVTM-use}) whose outputs are generated by the cross-modal encoder. Note that the cross-modal encoder in the P3E version is actually a three-layers BERT-base transformer blocks, whose setting follows UniVL \cite{luo2020univl}. Moreover, we find that the pre-training would hardly converge under the setting of employing the original BERT-Base encoder with 12 layers of blocks. As shown in Table \ref{ablation_ant} and Figure \ref{ablation_tensorboard}, we have several observations as follows: 

\begin{itemize}

	\item SNP is more \textbf{lightweight}. We count total trainable parameters in Table \ref{ablation_ant}, both SNP-R50 and SNP-PVT reduce more than 20$\%$ of parameters compared with their P3E version (about 200M $\rightarrow$ 160M), which verifies that SNP effectively simplifies the original model size.
 
	\item SNP is more \textbf{time-efficient} as it takes less time to reach a comparable performance. As illustrated in Figure \ref{ablation_tensorboard}, SNP needs fewer steps to achieve the same performance as P3E for both masking tasks and matching tasks. 
 
	\item SNP is \textbf{easier to train and converge}. As illustrated in Figure \ref{ablation_tensorboard}, both masking and matching losses converge faster when equipped with SNP than with P3E. Besides, SNP avoids possible risks brought by an improper parameter initialization that P3E needs for training the separate cross-modal encoder. Moreover, as can be seen in Table \ref{ablation_ant} (\textit{A1} \textit{v}.\textit{s}. \textit{A2}, \textit{A3} \textit{v}.\textit{s}. \textit{A4}), SNP also shows better fine-tuning performance on several downstream video-text datasets, proving that SNP is more powerful than P3E.

\end{itemize}

For the above three observations, SNP is a lighter, faster, and stronger pre-training architecture than P3E.


\subsection{Ablation Study of ${\textbf{S}}^{\textbf{3}}$}

As aforementioned, the proposed Significant Semantic Strengthening (${\rm \textbf{S}}^{\textbf{3}}$) strategy is model-agnostic, parameter-free, and could evidently promote the fine-tuning performance. We conduct several ablation study to verify these advantages.

Our proposed ${\textbf{S}}^{\textbf{3}}$ strategy includes a novel masking loss (MSSM) and a matching one (LVWM). Table \ref{ablation_sss} presents the experimental results of employing different pre-training objectives. We have several observations as follows: 

\begin{itemize}

	\item Our proposed MSSM is a more \textbf{powerful} masking loss than conventional MLM. As shown in Table \ref{ablation_sss} (\textit{B1} \textit{v}.\textit{s}. \textit{B2}, \textit{B5} \textit{v}.\textit{s}. \textit{B6}), masking significant semantics rather than other trivial ones could force the model to predict these clozes according to textual and visual cues, which benefits the cross-modal interaction and further promotes the performance.
 
	\item Our proposed LVWM is an \textbf{effective complementary} matching loss to GVTM. As shown in Table \ref{ablation_sss} (\textit{B1} \textit{v}.\textit{s}. \textit{B3}, \textit{B5} \textit{v}.\textit{s}. \textit{B7}), both sentence-level global representations (the ``[$cls$]'' token) and word-level local information (token lists of some significant semantics) are beneficial for understanding a given sentence, and combining both of them could further facilitate the cross-modal interaction.
 
	\item Our proposed MSSM and LVWM are \textbf{model-agnostic}. As shown in Table \ref{ablation_sss} (\textit{B1} \textit{v}.\textit{s}. \textit{B4}, \textit{B5} \textit{v}.\textit{s}. \textit{B8}, \textit{B9} \textit{v}.\textit{s}. \textit{B10}), both MSSM and LVWM are encoder-independent and could largely promote the fine-tuning performance.
 
	\item Notably, our proposed MSSM and LVWM are both \textbf{parameter-free} objectives. As shown in Eq.\ref{loss-mssm-t}, Eq.\ref{loss-mssm-m}, and Eq.\ref{loss-LVWM}, both MSSM and LVWM do not introduce new parameters during computation. Therefore, they would not heavily slow down the speed of pre-training.

\end{itemize}

For the above four observations, we believe that ${\textbf{S}}^{\textbf{3}}$ is a model-agnostic and implementation-friendly strategy, which could efficiently promote the performance.

\begin{table}
	\small
	\centering
	\begin{tabular}{p{1.2cm}<{\centering}|p{1.6cm}<{\centering}|p{2.6cm}<{\centering}|p{1.0cm}<{\centering}}
		\hline
		\multicolumn{1}{c|}{Model} &
		\multicolumn{1}{c|}{Chosen Num} &
		\multicolumn{1}{c|}{Retrieval (7K-1K)} &
		\multicolumn{1}{c}{VideoQA}
		\\
		\hline
		 &${N}_{L}$=1 & 26.2 / 54.2 / 66.3 & 41.90  \\
		SNP-S$^{3}$ &${N}_{L}$=2 & 25.5 / 53.2 / 66.8 & 41.76 \\
		PVT &${N}_{L}$=3 & \textbf{26.6} / \textbf{55.5} / \textbf{67.7} & \textbf{42.00} \\
		&${N}_{L}$=4 & 26.4 / 54.7 / 67.5 & 41.94 \\ 
		\hline
		
	\end{tabular}
    \caption{Parameter analysis of the number of the chosen significant token features (${N}_{L}$). All experiments are evaluated on two tasks (Retrieval on 7K-1K split and VideoQA) of the MSRVTT dataset under the backbone of SNP-S$^{3}$-PVT.}
\vspace{-0.2cm}
\label{para_NL}
\end{table}

\subsection{Parameter Analysis}

We conduct the parameter analysis towards the number of chosen significant token features (${N}_{L}$) of the LVWM loss. As shown in Table \ref{para_NL}, the fine-tuning performance on the MSRVTT dataset first increases when adding more significant token features in computing the LVWM loss, and it would reach the peak at ${N}_{L}=3$. It is probably due to the fact that the significant words (NOUNs, VERBs, and ADJECTIVEs) in one sentence are limited, so expanding the chosen number would have a performance upper bound.  

\subsection{Qualitative Analysis of ${\textbf{S}}^{\textbf{3}}$}
To get an intuitive perception of the advantages of ${\textbf{S}}^{\textbf{3}}$, we employ Grad-CAM \cite{selvaraju2017grad}, a widely-employed ``visual explanation'' toolkit, to visualize the attention location map of the last convolution layer in PVT. As shown in Figure \ref{visual_grad}, compared with SNP-PVT, the improved version SNP-S$^{3}$-PVT tends to emphasize those informative words, thus the latter would better model the cross-modal interaction. \textit{E}.\textit{g}., SNP-PVT wrongly lays its attention on the surroundings in Image 1, while SNP-S$^{3}$-PVT correctly emphasizes the object ``\textit{plane}'' and its action ``\textit{land}''. While in Image 2, SNP-S$^{3}$-PVT successfully recognizes the scene ``\textit{surfer-riding-waves}'', while SNP-PVT fails to do so.

\begin{table}
	\small

	\centering
	\begin{tabular}{p{0.4cm}<{\centering}|p{3.4cm}<{\centering}|p{2.0cm}<{\centering}|p{0.8cm}<{\centering}}
		\hline
		\multicolumn{1}{c|}{No.} &
		\multicolumn{1}{c|}{Losses of SNP-PVT} &
		\multicolumn{1}{c|}{TVR (7K-1K)} &
		\multicolumn{1}{c}{VQA}
		\\ \hline
		C1 & MLM-AM, GVTM & 26.8/52.6/64.2 & 41.44 
		\\ \hline
		C2 & MSSM, GVTM & \textbf{27.2}/\textbf{53.2}/\textbf{65.7} & \textbf{41.74}
		\\ \hline \hline
		C3 & MLM, GVTM, TACo-L2 & 24.3/52.0/64.1 & 41.25  
		\\ \hline
		C4 & MLM, GVTM, LVWM & \textbf{25.5}/\textbf{52.4}/\textbf{64.7} & \textbf{41.72}
		\\ \hline

	\end{tabular}
    \caption{Performance comparison of related Significant Elements Mining methods and our proposed Significant Semantic Strengthening (\textbf{S$^{3}$}) strategy. For masking tasks, we compare the Attended Masking (MLM-AM) in VIOLET \cite{fu2021violet} and our MSSM task. For matching tasks, we compare the maximum token-level contrastive loss (TACo-L2) in TACo \cite{yang2021taco} and our LVWM task. All the experiments are conducted under the SNP-PVT backbone and evaluated on the MSRVTT dataset.}
\vspace{-0.2cm}

\label{vwm_vs_taco}
\end{table}

\subsection{Performance Comparison of Related Methods}
\label{sec:taco}
In Section \ref{sec:related}, we have introduced some Significant Element Mining methods, including the Attended Masking (denoted as MLM-AM) strategy proposed by VIOLET \cite{fu2021violet} and the maximum token-level contrastive loss (denoted as TACo-L2) proposed by TACo \cite{yang2021taco}. To prove the superiority of our Significant Semantic Strengthening (\textbf{S$^{3}$}) strategy, we reproduce MLM-AM and TACo-L2 based on the SNP architecture. As illustrated in Table \ref{vwm_vs_taco}, for masking strategies, MSSM outperforms MLM-AM by 0.4/0.6/1.5 on R@1/5/10 of TVR and 0.30 on the Accuracy of VQA. While for matching strategies, LVWM outperforms TACo-L2 by 1.2/0.4/0.6 on R@1/5/10 of TVR and 0.47 on the Accuracy of VQA. One possible reason is that TACo-L2 only takes one token with the maximum similarity score into computation, which may also omit some local information compared with LVWM that employ multiple informative semantics to model the sentence at the word level. These comparisons prove that \textbf{S$^{3}$} is an effective strategy to promote the pre-training performance.


\section{Conclusion}

In this paper, we improve conventional video-text pre-training methods from two aspects. For the pre-training architecture, we propose Shared Network Pre-training (\textbf{SNP}), a novel paradigm that effectively absorbs the advantages of two mainstream pixel-level models and overcomes their shortcomings. For pre-training proxy tasks, we propose the Significant Semantic Strengthening (\textbf{S$^{3}$}) strategy to optimize masking and matching tasks for better cross-modal interaction. In the future, we plan to employ a shared encoder to embed visual, textual, and cross-modal information from raw video data; and design a more robust Significant Semantic Mining algorithm to promote the cross-modal interaction.

\newpage
{\small
\bibliographystyle{ieee_fullname}
\bibliography{egbib}
}

\end{document}